\documentclass[acmlarge]{acmart}

\AtBeginDocument{%
  }

\setcopyright{acmlicensed}
\copyrightyear{2018}
\acmYear{2018}
\acmDOI{XXXXXXX.XXXXXXX}
\acmConference[Conference acronym 'XX]{Make sure to enter the correct
  conference title from your rights confirmation email}{June 03--05,
  2018}{Woodstock, NY}
\acmISBN{978-1-4503-XXXX-X/2018/06}

\usepackage[inline]{enumitem} 
\usepackage{amsmath}
\usepackage{multirow}
\usepackage{subcaption}
\usepackage{bm}
\usepackage{bbding}
\usepackage{xcolor}
\definecolor{revisionblue}{RGB}{0,90,200}

\DeclareRobustCommand{\revision}[1]{%
  {\leavevmode\begingroup\color{revisionblue}#1\endgroup}%
}

\DeclareRobustCommand{\revision}[1]{#1}

\usepackage{pifont}

\usepackage{hyperref}

\theoremstyle{plain}
    \newtheorem{theorem}{Theorem}

\theoremstyle{definition} 
  
  \newtheorem{observation}{Observation}

\begin{document}

\title{Beyond KAN: Introducing KarSein for Adaptive High-Order Feature Interaction Modeling in CTR Prediction}


\author{Yunxiao Shi}
\affiliation{%
  \institution{University of Technology Sydney}
  \city{Sydney}
  \state{NSW}
  \country{Australia}
}

\author{Wujiang Xu}
\affiliation{%
  \institution{Rutgers University}
  \city{New Brunswick}
  \state{NJ}
  \country{USA}
}

\author{Haimin Zhang}
\affiliation{%
  \institution{University of Technology Sydney}
  \city{Sydney}
  \state{NSW}
  \country{Australia}
}

\author{Qiang Wu}
\affiliation{%
  \institution{University of Technology Sydney}
  \city{Sydney}
  \state{NSW}
  \country{Australia}
}

\author{Min Xu}
\affiliation{%
  \institution{University of Technology Sydney}
  \city{Sydney}
  \state{NSW}
  \country{Australia}
}








\renewcommand{\shortauthors}{Yunxiao et al.}

\begin{abstract}
Modeling high-order feature interactions is crucial for click-through rate (CTR) prediction, yet traditional approaches typically predefine a maximum interaction order and exhaustively enumerate feature combinations up to that order. This paradigm depends heavily on prior domain knowledge to delimit the interaction space and incurs substantial computational overhead. As a result, conventional CTR models face a persistent tension between enriching representations with complex high-order interactions and keeping computation tractable. To address this dual challenge, this study introduces the Kolmogorov-Arnold Represented Sparse Efficient Interaction Network (KarSein). Drawing inspiration from the learnable activation mechanism in the Kolmogorov-Arnold Network (KAN), KarSein leverages this mechanism to adaptively transform low-order basic features into high-order feature interactions, offering a novel approach to feature interaction modeling. KarSein extends the capabilities of KAN by introducing a more efficient architecture that significantly reduces computational costs while accommodating two-dimensional embedding vectors as feature inputs. Furthermore, it overcomes the limitation of KAN's its inability to spontaneously capture multiplicative relationships among features.
Extensive experiments highlight the superiority of KarSein, demonstrating its ability to surpass not only the vanilla implementation of KAN in CTR prediction tasks but also other baseline methods. Remarkably, KarSein achieves exceptional predictive accuracy while maintaining a highly compact parameter size and minimal computational overhead. Moreover, KarSein retains the key advantages of KAN, such as strong interpretability and structural sparsity. As the first systematic adaptation of KAN to CTR prediction, KarSein offers a practical, parameter-efficient, and interpretable alternative for modeling complex feature interactions in large-scale recommendation systems \footnote{The code is available at: https://github.com/Ancientshi/KarSein}. 
\end{abstract}



\received{20 February 2007}
\received[revised]{12 March 2009}
\received[accepted]{5 June 2009}

\acmSubmissionID{430}
\maketitle


\section{Introduction}

In the rapidly evolving landscape of digital advertising and recommendation systems, accurately predicting Click-Through Rate (CTR) has become critically important for optimizing user engagement and maximizing revenue potential \cite{DLRM, CTR_review, wang2025causal, gharibshah2021user,AdaGIN,Factor_Disentanglement,MEGG}. Central to effective CTR prediction is capturing the intricate relationships among contextual features in complex user–item interactions \cite{FRNet, Elevating,causal_inference_RS_survey,QNN_for_interaction}. Existing feature modeling paradigms predominantly fall into two categories: implicit feature interactions, which primarily use deep neural networks, graph neural networks, or incorporate sequential architectures (e.g., Transformer \cite{FuXi}, LSTM \cite{LSTM_ctr}) to automatically fuse highly expressive features \cite{DNN_CTR, WDL, FinalMLP, NCF, qu2016product}. Such features, fused by black‐box models, typically lack physical interpretability and are not explainable. In contrast, explicit feature interaction methods, which typically employ factorization‐based models \cite{DeepFM, xDeepFM, neural_FM} to manually construct multiplicative high‐order features. Empirical results in the CTR domain indicate that constructing explicit feature interactions is crucial for achieving optimal prediction accuracy; relying solely on implicit feature interactions yields suboptimal performance. The reason is that these complex high‐order multiplicative relationships are difficult to represent implicitly using DNNs, GNNs, or Transformer architectures \cite{AutoInt, rendle2020neural}.

\begin{figure}[h]
  \centering
    \includegraphics[width=3.8 in]{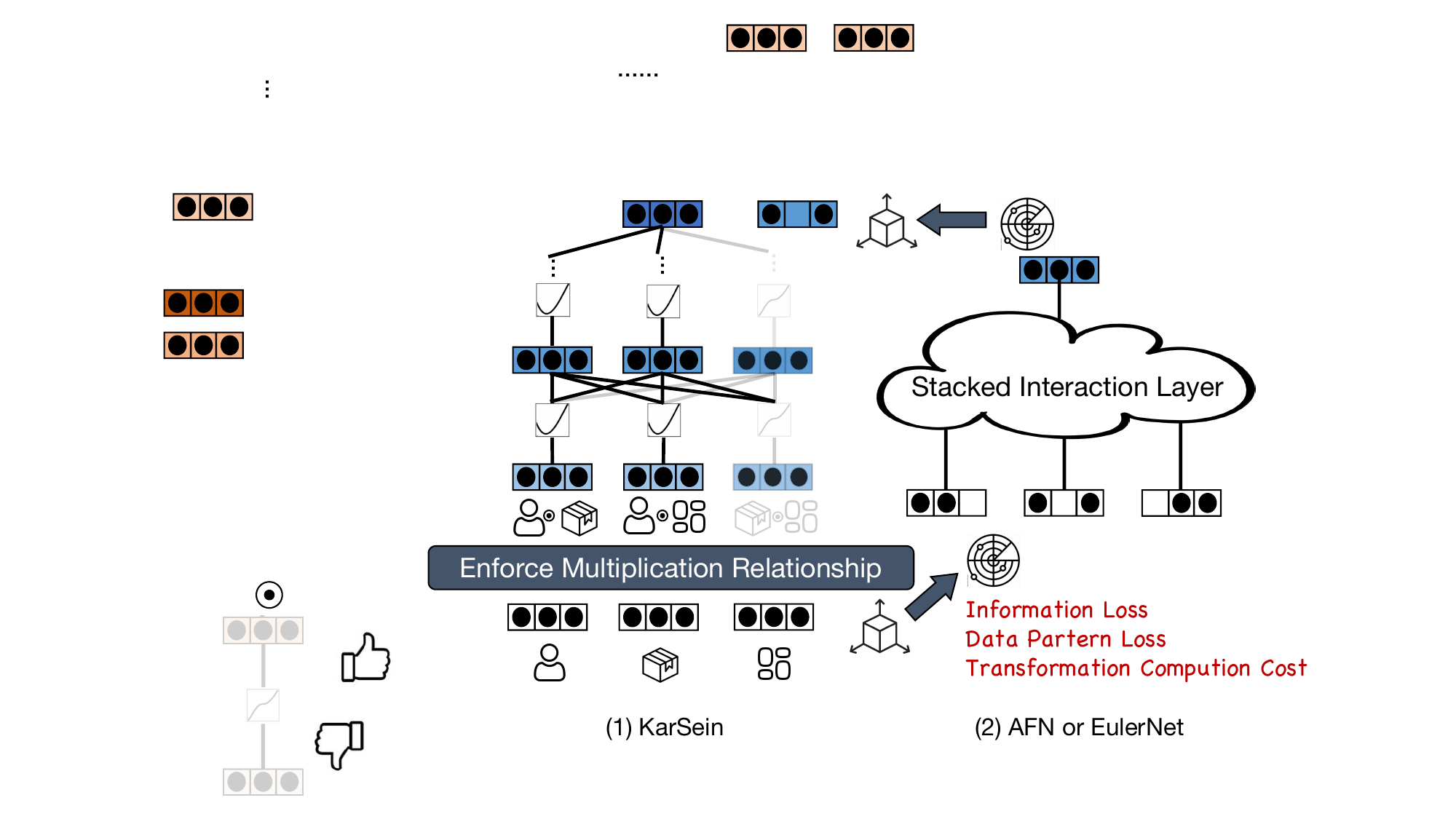}
   \caption{Comparison between KarSein and pioneer methods. KarSein models high order features interaction simply via activation without space transformation. It is more intuitively explainable and structural sparse.}
    \label{fig: architecture}
\end{figure}

Explicit feature interaction methods often struggle to capture high‐order features because the number of possible combinations grows exponentially with interaction order. As the feature set expands—often to hundreds of very sparse, high‐dimensional fields (e.g., user IDs or item IDs)—the total combinations quickly become astronomical. In practice, even performing fourth‐order interactions on just 30 raw features yields nearly 27,000 distinct combinations. This combinatorial explosion means that exhaustively computing and storing high‐order cross features becomes prohibitively expensive in both time and memory, making such approaches impractical for large‐scale, sparse real‐world data \cite{retrieve}. To address this issue,  AFN~\cite{AFN} and EulerNet~\cite{EulerNet} design novel feature transformations to adaptively learn high-order interactions. However, they require a preceding embedding space and face challenges such as information loss, numerical stability issues, computational overhead during spatial transformations, and various limitations. For instance, the logarithmic transformation applied in AFN \cite{AFN} requires strictly positive inputs and assumes rigid multiplicative or divisive relationships among the input features, whereas the polar coordinate transformation applied in EulerNet \cite{EulerNet} may be insufficient to model complex nonlinear feature entanglements due to its reliance on geometric decomposition. These drawbacks highlight a fundamental limitation of existing transformation‐based methods: their reliance on spatial reparameterization over embedded features often introduces instability and inefficiency, as illustrated in \autoref{fig: architecture} (2).

Motivated by these limitations, we propose modeling high-order feature interactions without spatial transformations by leveraging the learnable activation mechanism of Kolmogorov–Arnold Networks (KAN)~\cite{KAN}. KAN employs one-dimensional, spline-based activation functions directly embedded within the network edges, enabling accurate data fitting, intuitive interpretability, and structural sparsity \cite{KAN_adma,KAN_survey,KAN-AD}. Motivated by these strengths, we view KAN as a fundamentally novel approach to modeling feature interactions, capable of naturally transforming low-order univariate signals into high-order multivariate interactions. As illustrated in \autoref{fig: activation_visualiz}~(2), independent features are individually processed through learned activation functions and then combined. The resulting composite feature subsequently passes through another activation function. Assuming each activation corresponds to a third-degree polynomial, this two-step activation process effectively captures interactions up to the ninth degree. This characteristic underscores KAN’s ability to efficiently model complex, high-order interactions. By embedding transformations directly onto network edges rather than relying on explicit spatial embeddings or handcrafted mappings, KAN can alleviate some of the information loss, numerical instability, and computational overhead commonly encountered in traditional feature-transformation methods.

However, we observe several critical limitations when directly applying KAN to CTR prediction:  
(1) \textbf{Difficulty in Learning Multiplicative Feature Interactions (\autoref{obs: obs1}).} While theoretically capable, KAN struggles to autonomously capture the multiplicative patterns crucial for CTR tasks due to high sensitivity to initialization and regularization.
(2) \textbf{Prohibitive Computational Complexity (\autoref{obs: obs2}).} The significant computational complexity associated with KAN poses practical challenges, rendering it infeasible for deployment in real-world CTR prediction scenarios.
(3) \textbf{Incompatibility with Embedding Vectors (\autoref{obs: obs3}).} KAN inherently operates on scalar values, lacking native support for the vector-wise feature interactions essential for CTR recommendation systems.

To specifically address and overcome these challenges and adapt KAN effectively for CTR prediction tasks, we introduce a novel model named \textbf{Kolmogorov–Arnold Represented Sparse-Efficient Interaction Network (KarSein)}, which builds upon but significantly extends beyond KAN. 
\textbf{KarSein extends KAN to accommodate two-dimensional input features, treating embedding vectors as holistic units.} This allows distinct adaptive activations for entire vectors, enabling feature interactions at the vector-wise level—a capability absent in vanilla KAN’s bit-wise modeling (\autoref{sec:architectural_ext}).
\textbf{To facilitate the learning of multiplicative patterns, KarSein introduces an explicit pairwise-multiplication operation in its initial layers.} These built‑in “catalyst” terms guide the network’s subsequent learning trajectory, making it easier to discover and refine increasingly complex multiplicative representations. Consequently, KarSein quickly evolves from these foundational pairwise interactions to capture sophisticated, higher‑order feature interactions—capabilities that are critical for superior CTR‑prediction performance (\autoref{sec:pairwise_multiplication}).  \textbf{Furthermore, in contrast to KAN—which assigns multiple activation functions to each input feature—KarSein allocates only a single activation function per feature.} This key simplification substantially reduces computational overhead, resulting in a more streamlined and efficient architecture. 
(\autoref{sec:activation_trans})
Since KarSein is derived from KAN, it retains several of KAN’s superior performance with a relatively small number of parameters (\autoref{sec:overall_performance}), and strong interpretability (\autoref{sec:explanatin}).

In summary, our paper offers the following contributions.

(1) We present the first investigation into the applicability of Kolmogorov-Arnold Networks (KAN) for CTR prediction tasks. Our analysis reveals that KAN struggles to autonomously capture multiplicative feature interactions, does not support embedding-based features necessary for modeling vector-wise interactions, and suffers from prohibitive computational complexity. These shortcomings render KAN inadequate for CTR tasks, leading to poor predictive performance despite substantial computational costs.
    
(2) We propose KarSein, a novel CTR prediction model leveraging the Kolmogorov–Arnold representation theorem to effectively model complex feature interactions. KarSein significantly extends KAN by addressing its inherent limitations, specifically tailored for CTR tasks.

(3) KarSein retains key advantages of KAN, including parameter efficiency, interpretability, and superior data-fitting capability. Notably, it achieves optimal performance even with shallow and narrow network configurations, and its converged models transparently express feature modeling through symbolic regression. Crucially, due to the intrinsic suitability of the Kolmogorov–Arnold theorem for feature modeling, KarSein achieves state-of-the-art prediction performance in terms of AUC.

\revision{(4) Extensive experiments on four real-world datasets show that KarSein achieves state-of-the-art CTR prediction accuracy while using on average 50× fewer parameters than strong baselines (up to 660× fewer) and roughly one-sixth as many as KAN under matched hidden-layer settings. Guided by symbolic regression, we further show that KarSein supports interaction orders that grow exponentially with depth; even with the 3-layer configuration in our experiments, this already corresponds to very high-order interactions (e.g., at least 63rd order), making KarSein a highly compact yet expressive architecture for CTR prediction.}

\section{Preliminary}
\subsection{Problem Formulation for CTR}
Let $\mathcal{U}$ and $\mathcal{I}$ denote the sets of users and items, respectively. For a user-item pair $(u,i) \in \mathcal{U} \times \mathcal{I}$, we define $\mathbf{x}_{u,i} = [x_1, \ldots, x_m]$ as the feature vector capturing relevant attributes, including categorical (e.g., user and item IDs) and numerical (e.g., age) features.
The Click-Through Rate (CTR) prediction task aims to estimate $P(y = 1 \mid \mathbf{x}_{u,i})$, where $y \in \{0, 1\}$ indicates whether the user clicked on the item. Formally, we define:
\begin{equation}
\hat{y} = f(\mathbf{x}_{u,i}; \Theta)
\end{equation}
where $\hat{y}$ is the predicted click probability, $f$ is the predictive model, and $\Theta$ represents the model parameters.
Given a training dataset $\mathcal{D} = \{(u_j, i_j, \mathbf{x}_{(u_j,i_j)}, y_j)\}_{j=1}^N$ of $N$ instances, we optimize $\Theta$ to minimize the prediction error. A common approach is to minimize the log loss:
\begin{equation}
\mathcal{L}(\Theta) = -\frac{1}{N} \sum_{j=1}^N \left[ y_j \log \hat{y}_j + (1 - y_j) \log (1 - \hat{y}_j) \right]
\end{equation}

\subsection{Feature Interactions Modeling}
In recommendation systems, the model typically includes a trainable parameterized embedding layer, denoted as $E(\cdot)$. Given a categorical ID feature $x_j$, the model maps it into dense vectors within a low-dimensional latent space: $e_j = E(x_j) \in \mathbb{R}^{D}$, where $D$ is the embedding dimension. Feature interaction modeling is subsequently performed on these vectorized embeddings. Broadly, feature interaction paradigms fall into two categories: \textit{implicit} and \textit{explicit}, which are often complementary.

\subsubsection{Implicit Feature Interactions.}
Implicit feature interactions refer to latent dependencies between features that are not explicitly predefined but are instead automatically captured by the model architecture. Let $\mathbf{e}=e_1||e_2||\ldots||e_m$ denote the result of wide concatenated embedding vectors. Then let $\mathbf{W}$ be the linear transformation matrix applied to $\mathbf{e}$, and $\sigma$ be the non-linear activation function. For an $L$-layer DNN, implicit feature interaction modeling can be expressed as:

\begin{equation}
\text{MLP}(\mathbf{e}) = (\mathbf{W}_{L} \circ \sigma \circ \mathbf{W}_{L-1} \circ \sigma \circ \cdots \circ \mathbf{W}_2 \circ \sigma \circ \mathbf{W}_1) \mathbf{e}.
\end{equation}

Implicit methods, such as standard DNNs and DCN \cite{DNN_CTR,DCN,DCNv2}, typically learn interaction patterns at the \textit{bit-level}. This capability is underpinned by the Universal Approximation Theorem (UAT), which posits that a neural network with at least one hidden layer, containing a finite number of neurons, and using an appropriate activation function, can approximate any complex function.

\subsubsection{Explicit Feature Interactions.}
\label{sec:vector_interaction}
Explicit feature interactions typically operate at the \textit{vector-level}, often employing the Hadamard product (element-wise multiplication) among embedding vectors, as seen in factorization-based models \cite{FM,DeepFM,xDeepFM}. Given a feature set $\{x_1, x_2, \ldots, x_m\}$ with corresponding embedding vectors $\{e_1, e_2, \ldots, e_m\}$, the space of $k$-th order explicit interactions is defined as the set of all element-wise products of exactly $k$ vectors:
\begin{equation}
\label{eq:define_e}
\{ e_{n_1} \odot e_{n_2} \odot \cdots \odot e_{n_k} \mid n_1, n_2, \ldots, n_k \in \{1, 2, \ldots, m\} \}.
\end{equation}
We stack the $m^{k}$ combination vectors into a matrix $\mathbf{X}_{k}\in\mathbb{R}^{m^{k}\times D}$.
Each row, indexed by $(n_{1},\ldots,n_{k})$, represents a $k$-th order feature interaction
$f^{k}_{(n_{1},n_{2}\ldots,n_{k})}(x_{1},x_{2}\ldots,x_{m})$. Concatenating all interactions from the 1st to the $k$-th order yields a matrix $\mathbf{X}_{1\sim k} \in \mathbb{R}^{\ \sum_{i=1}^{k} m^{i} \; \times \; D}$, which is expressed as follows: 
\begin{equation}
\mathbf{X}_{1\sim k} = \begin{pmatrix} \mathbf{X}_1 \\ \mathbf{X}_2 \\ \vdots \\ \mathbf{X}_k \end{pmatrix},
\end{equation}
This brute-force construction explodes geometrically—the row count is
$
s \;=\; \sum_{i=1}^{k} m^{i},
$
so with just $m\!\approx\!100$ fields even $k=3$ already spawns more than a million interaction terms. Such a combinatorial surge quickly overwhelms memory and training time, rendering naïve enumeration untenable for modern, high-dimensional CTR data.

To maintain tractability, traditional models \cite{xDeepFM,KD-DAGFM,DCNv2} typically cap the interaction order $k\!\le\!3$ and prune redundant terms. However, this heuristic severely restricts the model's expressive power by excluding potentially significant higher-order relations. Consequently, recent advances have pivoted toward adaptive architectures capable of automatically discovering informative interactions of arbitrary order. These approaches aim to eliminate predefined order limits and exhaustive search, thereby balancing computational efficiency with predictive accuracy.

\subsection{A Kolmogorov–Arnold View of Feature Interactions}

\subsubsection{Kolmogorov-Arnold Representation Theorem}
Vladimir Arnold and Andrey Kolmogorov established a fundamental theoretical result concerning the representation of multivariate functions. Specifically, under the assumption that the function $f: [0,1]^m \rightarrow \mathbb{R}$, Kolmogorov-Arnold theorem states formally:

\begin{theorem}
\label{thm:KA}
Given a function $f: [0,1]^m \rightarrow \mathbb{R}$, there exist univariate functions $\phi_{q,p} : [0,1] \rightarrow \mathbb{R}$ and functions $\Phi_q : \mathbb{R} \rightarrow \mathbb{R}$ such that:
\begin{equation}
f(x_1, \ldots, x_m) = \sum_{q=1}^{2m+1} \Phi_q \left( \sum_{p=1}^{m} \phi_{q,p}(x_p) \right).
\end{equation}
\end{theorem}

This theorem implies a powerful structural result: any high-dimensional function can be expressed precisely as compositions and sums of a finite set of univariate functions. Consequently, the original problem of approximating or learning high-dimensional functions reduces significantly in complexity to learning a polynomial number of simpler one-dimensional functions. 

\noindent\textbf{Implications for CTR prediction.}
As a direct consequence of \autoref{thm:KA}, it is intuitive that \emph{any finite–order multivariate feature interaction is representable by compositions and sums of univariate functions}. This statement provides the theoretical bedrock: high-order multiplicative patterns can be captured without explicitly constructing an exponential cross-feature space; by learning the activation functions $\phi_{q,p}$ and $\Phi_q$, we can automatically uncover the required interactions. Each activations admit an analytic form, allowing \emph{symbolic regression} of the recovered interaction, which offers rare interpretability in CTR models, revealing which higher-order relations the predictor deems salient.

\subsubsection{Kolmogorov–Arnold Networks} 
\begin{figure}[h]
  \centering
    \includegraphics[width=3.0 in]{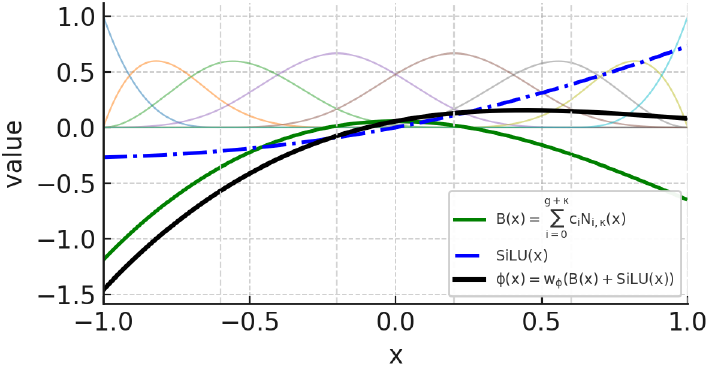}
   \caption{An activation function is parameterized as a B-spline.}
    \label{fig:activation}
\end{figure}

A central obstacle in applying \autoref{thm:KA} is that its one-dimensional component functions can be highly irregular—frequently non-smooth or even fractal—so they are ill-suited for direct use in machine-learning models. Kolmogorov–Arnold Networks (KAN) \cite{KAN}, however, \revision{present} an optimistic perspective and makes the insights practically usable. It employs highly flexible, learnable B-Spline curves as $ \phi(\cdot)$ for activation, and extends the theorem to neural networks with arbitrary widths and depths. Formally, for an activation function $ \phi(\cdot)$, let $ \text{Silu}(\cdot) $ denote the Silu activation, and $ \operatorname{B}(x) = \sum_{i=0}^{g+\kappa} c_i N_{i,\kappa}(x)$ represent the B-Spline curves activation withe grid size $g$ and order $\kappa$, where both $ w_{\phi} $ and $c_i$ are learnable parameters. The activation function $\phi(x)$ is defined as:
\begin{equation}
\phi(x) = w_{\phi} (\operatorname{B}(x) + \text{Silu}(x)),
\end{equation}
and the process of how $x$ is activated into $\phi(x)=0.2707x^3 - 0.7449x^2 + 0.4993x + 0.0561$ is illustrated in \autoref{fig:activation}, where $\kappa=3,g=5$,$
\mathbf{C}
  = \bigl[
    -1.189853,\;
    -0.826018,\; 
    -0.280551,\;
    0.075961,\;
    0.062688,\;
    -0.216423,\;
    -0.502392,\;
    -0.649853
   \bigr]
$, and $w_{\phi}=1$.

Let $\Phi_L$ is the activation function matrix corresponding to the L-th KAN layer. A general KAN network is a composition of $L$ layers, given an wide concatenated embedding vector $e$ as input, the output of KAN is:
\begin{equation}
\text{KAN}(e) = ( \Phi_{L} \circ  \Phi_{L-1} \circ \cdots \circ \Phi_2 \circ \Phi_1)e.
\end{equation}

\autoref{fig: architecture} (2) depicts a representative $mD\!-\!3\!-\!1$ Kolmogorov–Arnold Network, furnishing readers with a concrete point of reference as well as an intuitive visualisation of how activations propagate through the stacked layers.  In this canonical design, the input to the $i$-th layer is paired with as many learnable activation functions as there are nodes in layer $i\!+\!1$, a choice that quickly inflates the parameter budget. \autoref{sec:vanilla_kan_ctr} later shows that, for CTR prediction, this profusion of activations is largely superfluous and constitutes a major source of computational inefficiency.  Guided by this insight, \autoref{sec:activation_trans} introduces an optimised variant that assigns exactly one activation function to each feature. This streamlined configuration remains theoretically sound while markedly improving efficiency.

\section{Exploration Study}\label{sec:toy_feasibility}
\subsection{KAN for Learning Multiplicative Feature Interactions}
Having established that Kolmogorov–Arnold Networks (KANs) can express any finite-order feature interaction on theoretical grounds, we now turn to an empirical sanity check.  Our aim is to examine whether a KAN can actually learn simple quadratic interactions in practice—an essential prerequisite before tackling the far more intricate patterns found in real CTR data.

\subsubsection{Experimental setup}
We consider two raw scalar features, $a$ and $b$, whose second-order interactions comprise the monomials $a^{2}$, $b^{2}$, and the cross-term $ab$.  For each target function we hand-craft an \emph{idealised} KAN topology that is just expressive enough to fit the interaction in question: a \(2\!-\!1\) network for $a^{2}$, another \(2\!-\!1\) network for \(b^{2}\), and a slightly wider \(2\!-\!2\!-\!1\) network for the mixed product \(ab\).  Beyond these ideal blueprints, we also examine two practical initialisation strategies—simulating realistic deployments in which the architecture is selected heuristically and regularisation may or may not be applied. All models are trained with Adam (learning rate \(10^{-3}\)) until the root-mean-square error (RMSE) falls below \(0.05\). We document the number of optimization steps required to meet this criterion in \autoref{tab:three_settings}.

\begin{table}[t]
\caption{Steps required to achieve RMSE $\leq 0.05$ for three different KAN settings in fitting $a^2$, $b^2$, and $ab$. \textbackslash{} indicates the configuration is unable to reach the specified RMSE.}
\label{tab:three_settings}
\centering
\begin{tabular}{c|c|c|c|c|ccc} 
\hline
\multirow{2}{*}{\textbf{ID}} & \multicolumn{3}{c|}{\textbf{\textbf{KAN Layers}}} & \multirow{2}{*}{\textbf{Regularization}} & \multicolumn{3}{c}{\textbf{Steps to Converge}}                            \\ 
\cline{2-4}\cline{6-8}
                                     & \textbf{$a^2$} & \textbf{$b^2$} & \textbf{$ab$}   &                                          & \textbf{\textbf{$a^2$}} & \textbf{\textbf{$b^2$}} & \textbf{\textbf{$ab$}}  \\ 
\hline
1                                    & (2, 1)          & (2, 1)          & (2, 2, 1)       & 0.01                                     & 600                     & 600                     & 1200                    \\ 
\hline
2                                    & (2, 4, 1)      &(2, 4, 1)      & (2, 2, 4, 1)   & 0.01                                     & 350                     & 350                     & \textbackslash{}        \\ 
\hline
3                                    & (2, 4, 1)      & (2, 4, 1)      & (2, 2, 4, 1)   & 0.00                                     & 250                     & 250                     & 320                     \\
\hline
\end{tabular}
\end{table}

\begin{figure}[t]
  \centering
    \includegraphics[width=3.6 in]{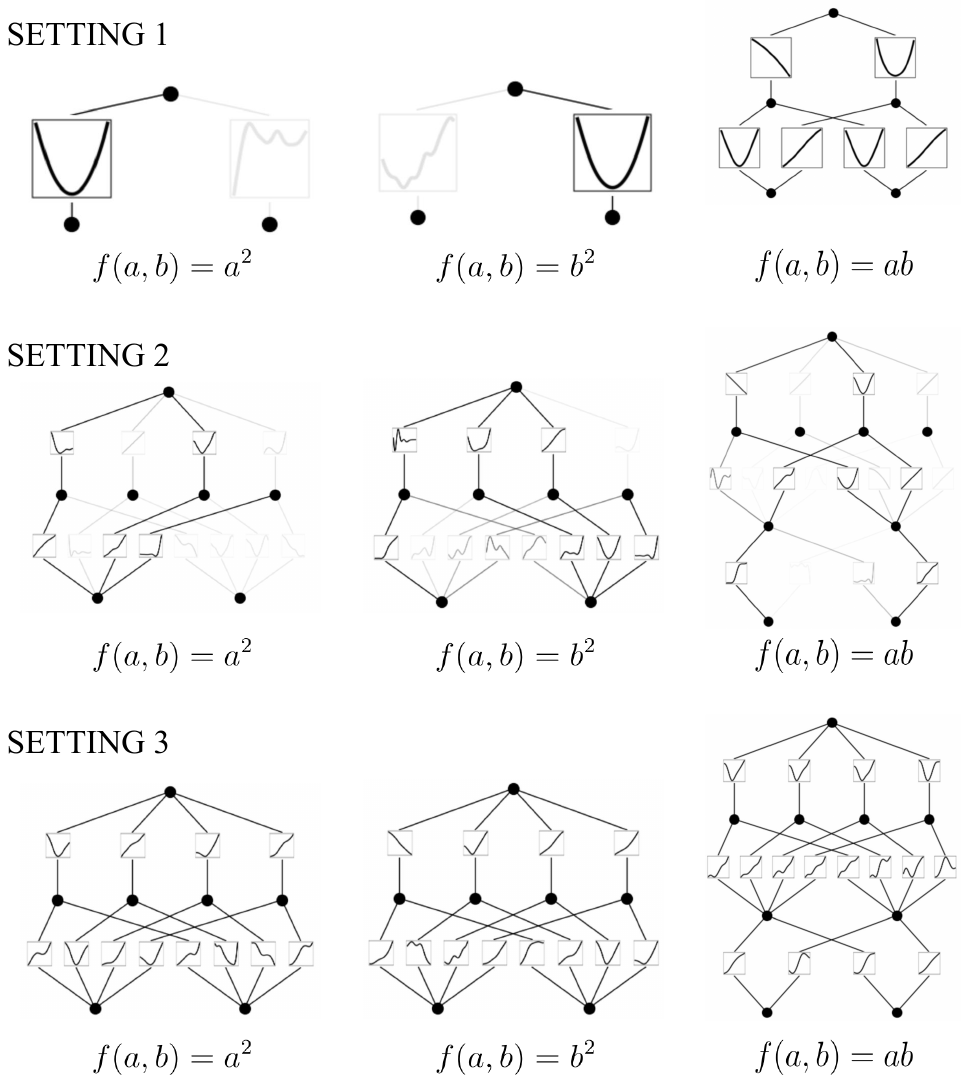}
   \caption{Visualization of KAN for fitting simple second-order feature interactions across three different settings.}
    \label{fig: KAN_feature_interaction}
\end{figure}

\subsubsection{Results} 
The experimental results confirm that, with an architecture precisely aligned to the target interaction (Setting~1), KAN converges rapidly and even yields an interpretable symbolic form, as visualised in \autoref{fig: KAN_feature_interaction}.  When the width is increased but regularisation is retained (Setting~2), the network still learns the independent squares but fails to approximate the cross term, suggesting that sparsity pressure prunes away critical paths before they can be strengthened by gradient updates.  Removing the penalty (Setting~3) restores the ability to model \(ab\); however, the resulting representations are dense and no longer correspond to clean symbolic decompositions.

\begin{observation} \label{obs: obs1}
KAN possesses ample capacity to fit multiplicative interactions, yet its success hinges on a delicate balance between architectural bias and regularisation strength. In realistic CTR scenarios—where neither ideal topologies nor manual pruning are available—the vanilla KAN is hard to discover multiplicative relations unaided.
\end{observation}

\begin{figure}[ht]
  \centering
    \includegraphics[width=2.8 in]{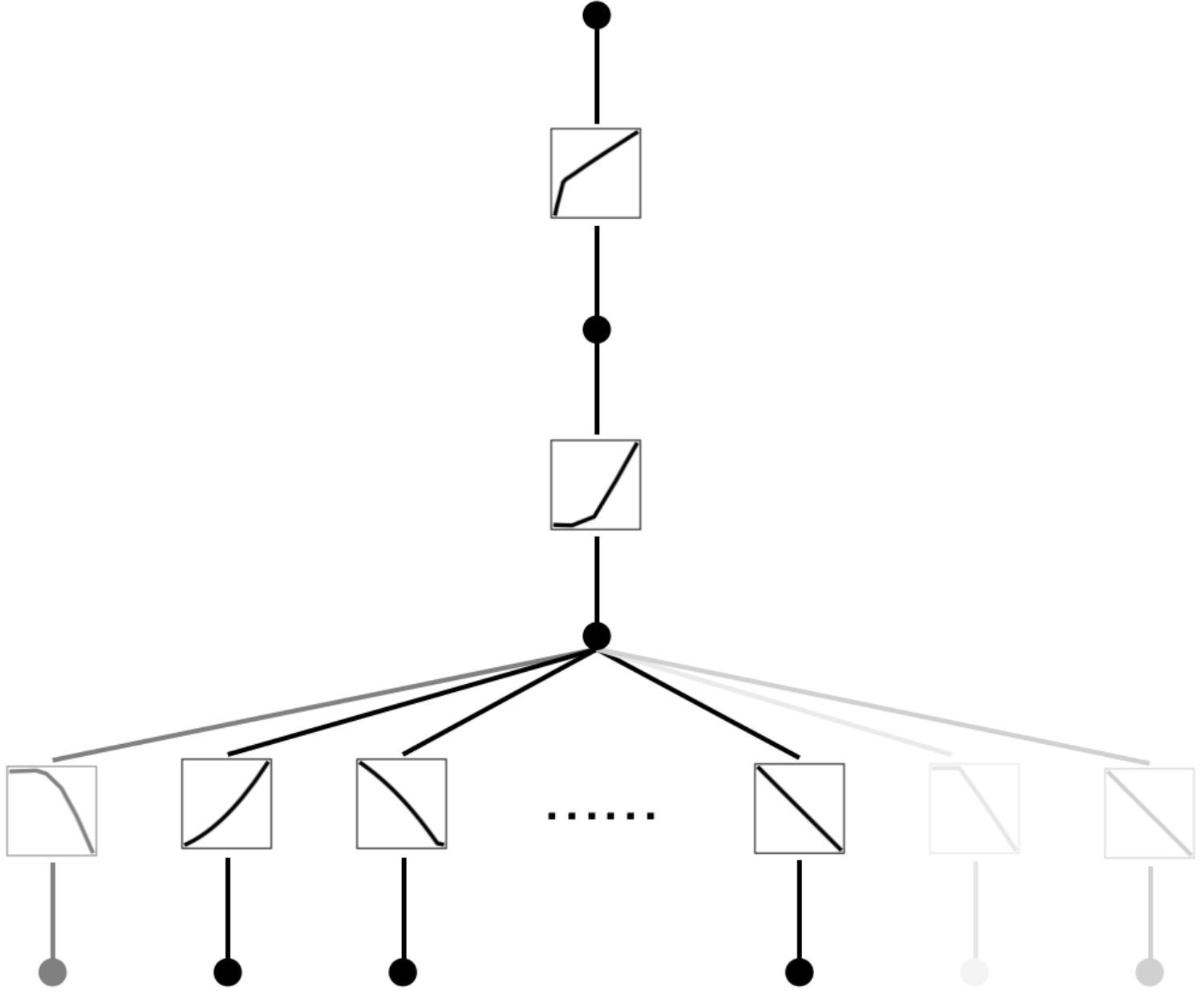}
   \caption{Visualization of KAN for CTR prediction.}
    \label{fig: KAN_CTR}
\end{figure}

\subsection{Vanilla KAN on MovieLens-1M}\label{sec:vanilla_kan_ctr}

To gauge how a straightforward Kolmogorov–Arnold Network behaves in a real CTR scenario, we evaluate a vanilla KAN on the widely used MovieLens-1M benchmark.  Hyper-parameters are fixed as follows: regularisation weight \(\lambda = 0.01\), learning rate \(10^{-2}\), batch size \(512\), B-spline grid size \(g = 3\), spline degree \(\kappa = 1\), embedding dimension \(D = 16\), and number of categorical fields \(m = 6\).  Training proceeds until the validation loss plateaus. The resulting model attains an AUC of 0.8273.  For reference, a conventional DNN with the same layer widths and neuron counts reaches 0.8403, underscoring a tangible performance gap in favour of the baseline. We subsequently prune every neuron whose incoming \emph{or} outgoing weight magnitude does not exceed \(0.003\).  As illustrated in \autoref{fig: KAN_CTR}, this procedure collapses the architecture from its initial \(96\!-\!64\!-\!64\!-\!1\) configuration to a remarkably compact \(96\!-\!1\!-\!1\!-\!1\) network—evidence of substantial parameter redundancy in the vanilla design.

\begin{observation}\label{obs: obs2}
Directly transplanting KAN into CTR prediction yields sub-optimal accuracy relative to an equally sized DNN. And allocating multiple activations per feature is unnecessary; one well-chosen activation per field is likely sufficient, and far more parameter-efficient.
\end{observation}

\begin{observation} \label{obs: obs3}
A further shortcoming of vanilla KAN is its \emph{inability to ingest embedding vectors directly}.
Because it processes features as independent scalars, it cannot model \emph{vector-wise interactions}—a capability that is crucial for modern CTR systems where each field is typically represented by a dense vector.  
\end{observation}

\section{Methodology}\label{sec:karsein_intro}
The empirical findings in \autoref{obs: obs1}--\ref{obs: obs3} make it clear that \emph{vanilla} Kolmogorov--Arnold Networks (KANs) fall short on three counts that are pivotal for large-scale CTR prediction: they neither capture multiplicative signals reliably, nor scale gracefully in computation, nor support modeling vector-wise feature interactions.  To bridge this gap, we propose \textbf{KarSein}—the \emph{Kolmogorov–Arnold Represented Sparse-Efficient Interaction Network}.  KarSein re-architects the original KAN layer into a lightweight \emph{KarSein Interaction Layer (KIL)} that accepts both scalar values and vector embeddings, thereby enabling both bit-wise and vector-wise feature modeling for solving issues in \autoref{obs: obs3}. KIL folds three mutually reinforcing operators: 
To directly remedy the deficiency highlighted in \autoref{obs: obs1}, we embed \emph{pairwise-multiplication operator} ($\mathcal{P}$) at the entrance of each KIL; by appending the products of the features, $\mathcal{P}$ serves as a catalytic scaffold that guides subsequent optimization toward discovering progressively richer multiplicative relationships (\autoref{sec:pairwise_multiplication}). Next, an \emph{activation-transformation operator} (\(\mathcal{A}\)) projects every feature onto a shared B-spline basis whose coefficients are feature-specific yet \emph{single} rather than multiple in KAN, therefore eliminating the redundant multi-activation design diagnosed in \autoref{obs: obs2} (\autoref{sec:activation_trans}). Finally, a \emph{linear-combination operator} (\(\mathcal{L}\)) mixes the activated signals into higher-order polynomial interactions (\autoref{sec:linear_comb}).

\subsection{Architectural Extension for Vector-Wise Feature Interactions}
\label{sec:architectural_ext}

Bit-wise (scalar) feature interactions trivially belong to the
Kolmogorov--Arnold (KA) function class.
We therefore focus on lifting the KA representation to vector-wise
(embedding) interactions. Let $\{e_1,\dots,e_m\}\subset[0,1]^D$ be embedding vectors for $m$ fields,
with coordinates $e_j^{(d)}$ for $d\in[D]:=\{1,\dots,D\}$.
Fix an interaction order $k\in\mathbb{N}$ and an index tuple
$(n_1,\dots,n_k)\in[m]^k$. Let
$
\mathbf e := (e_1,\dots,e_m)
$.
Consider the vector-wise feature interaction
\[
f^{k}_{\mathrm{vec}}(\mathbf e)
= e_{n_1}\odot\cdots\odot e_{n_k}\in\mathbb{R}^{D},
\]
whose $d$-th coordinate is
\[
f^{k}_{\mathrm{vec},d}(\mathbf e)
= \prod_{p=1}^{k} e_{n_p}^{(d)}.
\]
Then $f^{k}_{\mathrm{vec},d}$ admits a Kolmogorov--Arnold representation: for each $d\in[D]$ there exist univariate inner
functions $\{\phi^{(d)}_{q,p}\}$ and outer functions $\{\Phi^{(d)}_q\}$
such that
\begin{equation}
\label{eq:ka_vector}
f^{k}_{\mathrm{vec},d}(\mathbf e)=
\sum_{q=1}^{2k+1}\Phi^{(d)}_q\!\Bigl(\sum_{p=1}^{k}
        \phi^{(d)}_{q,p}\bigl(e_{n_p}^{(d)}\bigr)\Bigr).
\end{equation}
Stacking the $D$ scalar representations in \eqref{eq:ka_vector}
recovers the vector-wise feature interaction $f^{k}_{\mathrm{vec}}$. 

Since the coordinates of an embedding vector are usually interpreted as homogeneous “dimensions of meaning’’ in a shared latent space, there is no intrinsic reason to assign separate KA inner/outer functions to each coordinate. Classic embedding methods such as word2vec and GloVe already exploit this isotropy: semantics are encoded in the geometry of the whole vector, and downstream operations (e.g., dot products, linear maps) treat all dimensions symmetrically rather than giving each coordinate its own parametrization \cite{word2vec}. Likewise, modern embedding-based architectures in recommendation and NLP (e.g., DeepFM, xDeepFM’s vector-wise CIN module, and Transformer/BERT encoders) apply shared nonlinear transformations to entire fields or token embeddings \cite{DeepFM,xDeepFM,bert}, implicitly tying parameters across dimensions when modeling feature interactions. Given that embedding coordinates are thus exchangeable within a common semantic space, we can set
$$
\phi^{(1)}_{q,p}=\dots=\phi^{(D)}_{q,p}=: \phi_{q,p},
\qquad
\Phi^{(1)}_{q}=\dots=\Phi^{(D)}_{q}=: \Phi_{q}.
$$
This parameter tying preserves the KA representation while reducing the parameter footprint and improving efficiency.

In the following sections, we therefore take the embedding vector as the
basic unit of granularity, and implement vector-wise feature interactions
in KarSein with shared activation functions across
embedding dimensions originating from the same field.

\subsection{Architecture Overview}
 
The KarSein model is constituted by multiple stacked KarSein Interaction Layers. Each layer takes a set of embedding vectors as input and outputs high-order features. For the $L$-th KarSein Interaction Layer, the input dimension is $H_{L-1}$, and the output dimension is $H_{L}$. Here, $L \in [1, \mathcal{T}]$, and $\mathcal{T}$ denotes the depth of the stacked KarSein interaction layers. For the $L$-th layer, let $\mathbf{X}^{L-1} \in \mathbb{R}^{H_{L-1} \times D}$ represent the input matrix, and $\mathbf{X}^{L} \in \mathbb{R}^{H_{L} \times D}$ represent the output matrix.
Specifically, we define $H_{0} = m$, and $\mathbf{X}^{0} \in \mathbb{R}^{m \times D}$ as the matrix formed by stacking $e_{1}, e_{2}, \ldots, e_{m}$. Additionally, we define $H_{\mathcal{T}} = 1$ because the final layer is designed to have a single output neuron, ultimately modeling a highly intricate high-order feature interaction. For the \(L\)-th KarSein Interaction Layer (\(L=1,\dots,\mathcal{T}\)) we define
\begin{equation}
\Psi_{L} \;=\; \mathcal{L}_{L}\circ\mathcal{A}_{L}\circ\mathcal{P}_{L},
\end{equation}
where
\begin{equation}
\mathcal{P}_{L}:\mathbb{R}^{H_{L-1}\times D}\longrightarrow\mathbb{R}^{\tilde H_{L-1}\times D},\quad
\mathcal{A}_{L}:\mathbb{R}^{\tilde H_{L-1}\times D}\longrightarrow\mathbb{R}^{\tilde H_{L-1}\times D},\quad
\mathcal{L}_{L}:\mathbb{R}^{\tilde H_{L-1}\times D}\longrightarrow\mathbb{R}^{H_{L}\times D}.
\end{equation}
Stacking the \(L\) layers yields the overall KarSein mapping
\begin{equation}
\operatorname{KarSein}(\mathbf{X}^{0})\;=\;
\bigl(\Psi_{L}\circ\Psi_{L-1}\circ\cdots\circ\Psi_{1}\bigr)(\mathbf{X}^{0}).
\end{equation}
A schematic comparison between KarSein and the original KAN architecture is provided in \autoref{fig: architecture}; KarSein retains KAN’s interpretability and sparsity while directly overcoming its three CTR-specific shortcomings.

\begin{figure*}[tb!]
  \centering
    \includegraphics[width=6.0 in]{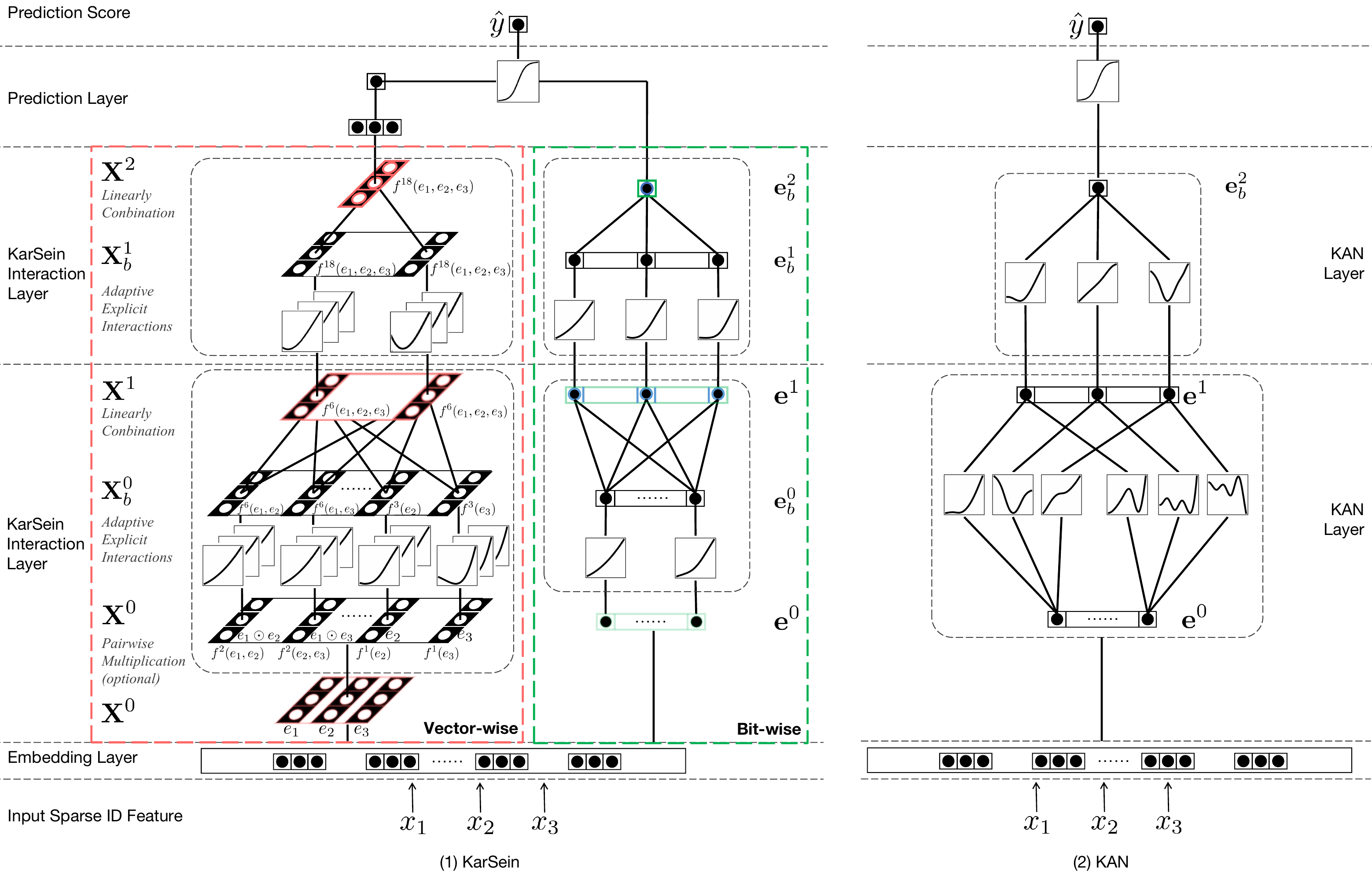}

    \caption{ \revision{ 
    Network Architectures and Visualization of Feature Interaction Modeling in KarSein and KAN. 
    Subfigure (1) illustrates KarSein, where the left branch is the vector-wise architecture for modeling explicit feature interactions and the right branch is the bit-wise architecture for modeling implicit feature interactions. Subfigure (2) shows the KAN baseline. In this toy schematic, the KarSein-explicit branch is configured with an $m$–2–1 architecture, while KarSein-implicit and KAN both use an $mD$–3–1 architecture. For KarSein, we color explicit/implicit feature interactions with a light-to-dark red/green gradient to indicate progressively higher orders. We further annotate the order of explicit feature interactions from 1 order at the input to 18 order at the output, computed by \autoref{eq:order_calculation} as $2 \times 3^2 = 18$. 
    }}
    \label{fig: architecture}
\end{figure*}

\subsection{KarSein Interaction Layer}
\label{sec:karsein_interaction_layer}

\subsubsection{Pairwise Multiplication}\label{sec:pairwise_multiplication}

To remedy the limitation highlighted in Observation~\ref{obs: obs1}—namely, vanilla KAN’s difficulty in autonomously uncovering multiplicative dependencies—we introduce a lightweight \emph{pairwise-multiplication operator}.  Given the hidden representation of the $(L\!-\!1)$-st layer, $\mathbf{X}^{L-1}\!\in\!\mathbb{R}^{H_{L-1}\times D}$, and the original embedding matrix $\mathbf{X}^{0}\!\in\!\mathbb{R}^{m\times D}$, this operator $\mathcal P_L$ \revision{updates} $\mathbf X^{L-1}$:
\begin{equation}
\mathbf X^{L-1}\;\gets  \;
\Bigl[\;
      \mathbf X^{L-1}\;
      \Big\|\;
      \bigl(\mathbf X^{L-1}\odot_{\text{row}}\mathbf X^{0}\bigr)
\Bigr]
\in\mathbb{R}^{\tilde H_{L-1}\times D},
\qquad
\tilde H_{L-1}=H_{L-1}(m+1),
\end{equation}
where $\odot_{\text{row}}$ indicates that for each row we first form the outer products, apply element-wise Hadamard products, and then concatenate the resulting cross features along the row dimension.

In mainstream CTR architectures (e.g., DCN \cite{DCN}, DCN-v2 \cite{DCNv2}, xDeepFM \cite{xDeepFM}), similar feature cross operations are enabled at every layer to explicitly enumerate ever-larger cross-feature spaces that plain DNNs struggle to capture. Our design philosophy diverges from traditional methods—motivation, intention, and practice.  We \emph{do not} seek to hand-craft interactions at every depth. Instead, $\mathcal P_L$ serves as a \emph{catalyst}: activating it only in the \emph{initial} interaction layers injects a strong multiplicative bias that subsequent spline activations can amplify into higher-order products on their own.

\subsubsection{Activation Transformation}\label{sec:activation_trans}

Let the B–spline basis of grid size~$g$ and order~$\kappa$ be  
\begin{equation}
\mathbf N_{\kappa}
=\bigl[N_{1,\kappa},\,N_{2,\kappa},\,\dots,\,N_{g+\kappa,\kappa}\bigr].
\end{equation}
For $L$-th layer's matrix for activation is $\mathbf{X}^{L-1} \in \mathbb{R}^{\tilde H_{L-1} \times D}$.
First, every row of $\mathbf X^{L-1}$ is projected onto the
$g+\kappa$ spline bases:
\begin{equation}
\mathbf X^{L-1}_{\text{basis}}
=\mathbf N_{\kappa}\!\bigl(\mathbf X^{L-1}\bigr)
\;\;\in\;\;
\mathbb R^{\tilde H_{L-1}\times D\times(g+\kappa)}.
\end{equation}
For every feature we have spline coefficients, we use $\mathbf C^{L-1}$ denotes it:
\begin{equation}
\mathbf C^{L-1}\in\mathbb R^{\tilde H_{L-1}\times 1\times(g+\kappa)},
\end{equation}
for every $\mathbf X^{L-1}_{\text{basis}}[i,:,:]$ it is corresponded with coefficient $\mathbf C^{L-1}[i,:,:]$.
Applying these coefficients yields the activated representation

\begin{equation}
\mathbf{X}^{L-1}_b = \begin{bmatrix} \mathbf{X}^{L-1}_{\text{basis}}[1,:, :] \mathbf{C}^{L-1}[1, :, :]^T \\ \vdots \\ \mathbf{X}^{L-1}_{\text{basis}}[H_{L-1}, :, :] \mathbf{C}^{L-1}[H_{L-1}, :, :]^T \end{bmatrix} 
\in\mathbb R^{\tilde H_{L-1}\times D}.
\end{equation}

\paragraph{Relation to KAN}
In a KAN layer, each input feature is paired with
$H_L$ distinct coefficient tensors, producing $H_L$ activation
functions whose outputs are then summed through an additional
learnable weight matrix.  We observe that the two successive linear
operations (spline combination and weighted summation) can be
merged into a single linear combination without loss of
expressiveness.  Hence KarSein retains only one coefficient tensor
$\mathbf C^{L-1}$ (a single activation function per feature),
eliminating redundant parameters and computation while remaining
faithful to the transformation principle and to the empirical insight
of \autoref{obs: obs2}.

\subsubsection{Linearly Combination} \label{sec:linear_comb}
We define the weight matrix $\mathbf{W}^{L-1}_b \in \mathbb{R}^{H_{L} \times \tilde H_{L-1}}$. To model the feature interactions, we perform a linear combination of the activated embedding vectors, represented as 
\begin{equation}
\mathbf{W}^{L-1}_b \mathbf{X}^{L-1}_b.
\end{equation}
To enhance the expressiveness of the model, we introduce an additional residual connection. Specifically, we apply the $\text{SiLU}(\cdot)$ activation function to $\mathbf{X}$ and define another weight matrix $\mathbf{W}^{L-1}_s \in \mathbb{R}^{H_{L} \times \tilde H_{L-1}}$ to perform a linear transformation on the activated embeddings, expressed as 
\begin{equation}
\mathbf{W}^{L-1}_s \text{SiLU}(\mathbf{X}^{L-1}).
\end{equation}
The final output features are then given by the following formulation:

\begin{equation}
  \mathbf{X}^{L}=\mathbf{W}^{L-1}_b \mathbf{X}^{L-1}_b + \mathbf{W}^{L-1}_s \text{SiLu}( \mathbf{X}^{L-1} )
\end{equation}

\subsection{Integrating Implicit Interactions}
We integrate implicit interactions, which are focused on bit-wise level feature interactions. We employ a parallel network architecture that separates the modeling of vector-wise and bit-wise interactions, with both networks sharing the same embedding layer. This classic parallel architecture has been extensively utilized in prior works \cite{DeepFM,xDeepFM,DCNv2,FinalMLP}.

For bit-wise interactions, we input a wide concatenated vector $\mathbf{e} \in \mathbb{R}^{mD}$   into the first layer. The total number of stacked KarSein interaction layers is $\mathcal{T}$, the final layer is designed to have a single neuron.   The output of the $\mathcal{T}$-th layer is denoted as  $\mathbf{e}^{\mathcal{T}} \in \mathbb{R}^{1}$.

\subsection{CTR Prediction}
The final outputs from the KarSein architecture for explicit feature interaction $\mathbf{X}^{\mathcal{T}}$, and the outputs from the KarSein architecture for implicit feature interaction $\mathbf{e}^{\mathcal{T}}$, are combined for binary classification in CTR prediction. The predicted probability, represented by $\hat{y}$, is calculated as follows:

\revision{
\begin{equation}
\hat{y}
= \sigma\!\big(\mathbf{X}^{\mathcal{T}} \mathbf{W}_o + \mathbf{e}^{\mathcal{T}}\big)
= \frac{1}{1 + \exp\!\big(-(\mathbf{X}^{\mathcal{T}} \mathbf{W}_o + \mathbf{e}^{\mathcal{T}})\big)} ,
\end{equation}
}

where $\mathbf{W}_o$ represents the regression parameters.

\subsection{Training with Sparsity}
\label{sec: sparsity}
The KAN network exhibits sparsity by applying L1 regularization to the parameters of the activation functions and entropy regularization to the activated values. Our model inherits this feature with enhanced efficiency. Instead of applying L1 regularization to the activation functions' parameters and entropy regularization to the post-activated values of intermediate input-output features, we incorporate L1 and entropy regularization into the KarSein interaction layer's linear combination step to eliminate redundant hidden neurons.

Specifically, for the $L$-th layer of the KarSein interaction layer, we apply L1 regularization to $\mathbf{W}^{L-1}_b$ and $\mathbf{W}^{L-1}_s$ with the regularization parameter $\lambda_1$. The L1 regularization term is computed as follows:

\begin{equation}
\lambda_1  ||\mathbf{W}^{L-1}_b||_1 + \lambda_1  ||\mathbf{W}^{L-1}_s||_1
\end{equation}

Next, we compute the entropy regularization term for $\mathbf{W}^{L-1}_b$ and $\mathbf{W}^{L-1}_s$ with the regularization parameter $\lambda_2$. The computation is as follows:

\begin{equation}
\lambda_2 \, \mathcal{H} \left( \frac{\mathbf{W}^{L-1}_b}{  ||\mathbf{W}^{L-1}_b||_1  } \right) + \lambda_2 \, \mathcal{H} \left( \frac{\mathbf{W}^{L-1}_s}{  ||\mathbf{W}^{L-1}_s||_1   } \right)
\end{equation}

where $\mathcal{H}(\cdot)$ denotes the entropy calculation. The total training objective is given by:

\begin{align*}
\mathcal{L}_{\text{total}} = \mathcal{L}_{\text{pred}} + \lambda_1 \sum_{L=1}^{\mathcal{T}} \left( \|\mathbf{W}^{L-1}_b\|_1 + \|\mathbf{W}^{L-1}_s\|_1 \right) + \\
\lambda_2 \sum_{L=1}^{\mathcal{T}} \left( \mathcal{H} \left( \frac{\mathbf{W}^{L-1}_b}{\|\mathbf{W}^{L-1}_b\|_1} \right) + \mathcal{H} \left( \frac{\mathbf{W}^{L-1}_s}{\|\mathbf{W}^{L-1}_s\|_1} \right) \right)
\end{align*}

\subsection{FLOPs Analysis}
\label{sec: KAN_CTR_Analysis}
Let $\mathcal{T}$ \revision{be} the depth of both KarSein-explicit and KarSein-implicit layers, and $H$ as the number of hidden neurons per layer. Additionally, $K$ represents the number of logarithmic neurons in AFN+ \cite{AFN}, which significantly exceeds $mD$. The parameter $n$ denotes the number of order vectors in EulerNet, is typically set to $m$ in practical applications. Then we present a comparative analysis, in terms of floating-point operations (FLOPs), among the proposed KarSein model, KAN, and other baselines, as outlined in \autoref{tab: table_comparison}.

We compare DNN, KAN, and KarSein-implicit in the context of implicit feature interaction modeling. DNN demonstrates higher efficiency than KAN; however, KAN compensates by requiring significantly fewer hidden neurons $ H $ and shallower layers $ \mathcal{T} $. The proposed KarSein-implicit method, with parameters $ g $ and $ \kappa $ much smaller than $ H $, achieves computational efficiency comparable to DNN while retaining KAN's advantage of smaller $ H $ and $ \mathcal{T} $.

In comparison of AFN+, EulerNet, and KarSein-explicit models which are designed to adaptively learn high-order features, AFN+ and EulerNet involve embedding space transformation. In contrast, KarSein-explicit performs feature interactions directly within the original space. Additionally, KarSein-explicit benefits from KAN’s structural sparsity feature, resulting in better explainability and feature interactions with less redundancy. Notably, KarSein-explicit exhibits computational complexity independent of $ D $, significantly outperforming AFN+ and EulerNet, and even surpassing DNN due to its smaller $ \mathcal{T} $ and $ H $ (often set to $ m \sim m^2 $).

\begin{table*}[h]
\small
\centering
\caption{FLOPs comparison of different methods.}
\label{tab: table_comparison}

\begin{tabular}{lccccc} 
\hline
\multirow{2}{*}{Methods} & \multicolumn{3}{c}{Feature Interaction}          & \multirow{2}{*}{\begin{tabular}[c]{@{}c@{}}Feature De-redundancy\end{tabular}} & \multirow{2}{*}{FLOPs}                         \\ 
\cline{2-4}
                         & High-order & Adaptive & Original Space &                                                                                                   &                                                     \\ 
\hline
DNN~\cite{DNN_CTR}                       & \textcolor{purple}{\XSolidBrush}          & \textcolor{purple}{\XSolidBrush}        & \textcolor{teal}{\CheckmarkBold}                        & \textcolor{purple}{\XSolidBrush}                                                       & $O(H(mD + H\mathcal{T})) $                           \\
KAN~\cite{KAN}                      & \textcolor{purple}{\XSolidBrush}          & \textcolor{purple}{\XSolidBrush}        & \textcolor{teal}{\CheckmarkBold}                        & \textcolor{teal}{\CheckmarkBold}                                                   & $O(H(mD+H\mathcal{T})(g + \kappa) $  \\

KarSein-implicit                  &\textcolor{purple}{\XSolidBrush}             & \textcolor{purple}{\XSolidBrush}          & \textcolor{teal}{\CheckmarkBold}                        & \textcolor{teal}{\CheckmarkBold}                                                       &           $O((mD+H\mathcal{T})(g+\kappa+H))$                                          \\
\hline
AFN+~\cite{AFN}                     & \textcolor{teal}{\CheckmarkBold}          & \textcolor{teal}{\CheckmarkBold}        & \textcolor{purple}{\XSolidBrush}                        & \textcolor{purple}{\XSolidBrush}                                                       &                  $O(mDK+KHD+mHD+H^2\mathcal{T})$                                   \\
EulerNet~\cite{EulerNet}                 & \textcolor{teal}{\CheckmarkBold}          & \textcolor{teal}{\CheckmarkBold}        & \textcolor{purple}{\XSolidBrush}                        & \textcolor{purple}{\XSolidBrush}                                                       & $O(mnD^2+n^2D^2\mathcal{T})) $                     \\
KarSein-explicit                  & \textcolor{teal}{\CheckmarkBold}          & \textcolor{teal}{\CheckmarkBold}        & \textcolor{teal}{\CheckmarkBold}                        & \textcolor{teal}{\CheckmarkBold}                                                       &              $O((m^2+H\mathcal{T})(g+\kappa+H))$                                       \\

\hline
\end{tabular}
\end{table*}

\section{Experiments}
\label{sec: exp}
In this section, we are to address these research questions: 
\begin{itemize}
\item \textbf{RQ1:} How does KarSein model perform compared to other state-of-the-art methods for CTR prediction?

\item \textbf{RQ2:} How do the explicit and implicit components of the KarSein model perform individually in prediction? 

\item \textbf{RQ3:} \revision{How do different hyperparameter configurations affect the model's performance?}

\item \textbf{RQ4:} \revision{What orders of feature interactions does KarSein automatically learn, and how do these interactions manifest as interpretable symbolic expressions?}

\item \textbf{RQ5:} \revision{How effectively does KarSein distinguish important from redundant features, and what global structural sparsity pattern emerges in the learned network?
}
\end{itemize}

\subsection{Experiment Setups}
\subsubsection{Datasets}
We conduct experiments on three datasets, including MovieLens 1M \footnote{https://grouplens.org/datasets/movielens}, Douban Movie \footnote{https://github.com/Yaveng/FIRE/tree/main/dataset}, Avazu \footnote{https://paperswithcode.com/dataset/click-through-rate-prediction-avazu}, and Criteo \footnote{http://labs.criteo.com/2014/02/kaggle-display-advertising-challenge-dataset}, which have been utilized in previous studies \cite{AFN, EulerNet, DCNv2, DeepFM, BARS}. \revision{For the MovieLens-1M and Douban Movie datasets, we cast the original 1–5 explicit ratings into an implicit-feedback / CTR prediction setup: ratings 4–5 are treated as clear positive feedback (label 1), ratings 1–2 as clear negative feedback (label 0), and rating 3 as ambiguous and therefore excluded. This preprocessing protocol follows prior work such as DCNv2 \cite{DCNv2} and AutoInt \cite{AutoInt}.}

\begin{itemize}
\item  \textbf{MovieLens 1M}: MovieLens 1M is a widely used benchmark in the field of recommender systems. It consists of 1 million movie ratings provided by 6,000 users on 4,000 movies. Each rating ranges from 1 to 5 for rating prediction.

\item  \textbf{Douban Movie}: The Douban Movie dataset consists of 1 million movie ratings, which are collected from the Douban website and range between 1 to 5 for rating prediction. The dataset includes data from 10,000 users and 10,000 movies spanning the years 2008 to 2019.

\item  \textbf{Avazu}: The Avazu dataset contains 40.4 million ad-impression logs across 22 categorical fields collected over ten days, serving as a large-scale benchmark for CTR prediction.

\item  \textbf{Criteo}: The Criteo dataset comprises user logs collected over a period of 7 days. It contains 45 million examples and 39 features, including 26 categorical feature fields and 13 numerical feature fields. We discretize each numerical value using a logarithmic discretization method.
\end{itemize}

\subsubsection{Baseline Methods}
To rigorously evaluate the effectiveness of our method, we benchmark it
against a diverse collection of state-of-the-art baselines that span the
full spectrum of feature–interaction modeling strategies in CTR
prediction.  For clarity, we categorise these baselines into four
non-overlapping groups: (1) Methods only have implicit feature interactions, i.e., DNN \cite{DNN_CTR}, KAN \cite{KAN}, Wide \& Deep \cite{WDL}, DCNV2 \cite{DCNv2}, and FinalMLP \cite{FinalMLP}. (2) Methods have implicit feature interactions, and explicit feature interactions with predefined order, i.e., DeepFM \cite{DeepFM}, xDeepFM \cite{xDeepFM}. (3) Methods have implicit feature interactions, and explicit feature interactions with adaptive order, i.e., AFN+ \cite{AFN}, EulerNet \cite{EulerNet}. (4) Feature interactions built on GNN, i.e., FiGNN \cite{FiGNN}. The introduction of each baseline is as follows:

\begin{itemize}
    \item \textbf{DNN} \cite{DNN_CTR} is a straightforward model based on deep-stacked MLP architectures, which applies a fully-connected network after the concatenation of feature embeddings for CTR prediction.
    \item \textbf{KAN} \cite{KAN} offers a promising alternative to MLPs. In our approach, we replace MLPs with KAN, allowing concatenated feature embeddings to be processed through KAN for CTR prediction.
    \item \textbf{Wide \& Deep} \cite{WDL} combines a linear model for memorization of feature interactions with a DNN for generalization to capture high-order interactions.
    \item \textbf{DCNV2} \cite{DCNv2} uses the kernel product of concatenated feature vectors to model high-order interactions and integrates a DNN for implicit interactions.
    \item \textbf{DeepFM} \cite{DeepFM} combines FM to capture second-order interactions with DNN to model high-order interactions.
    \item \textbf{xDeepFM} \cite{xDeepFM} encodes high-order interactions into multiple feature maps and combines them with a DNN to model implicit interactions.
    \item \textbf{AFN+} \cite{AFN} AFN encodes features into a logarithmic space to adaptively learn arbitrary-order feature interactions, AFN+ additionally uses a DNN for implicit interactions, further improving the model capacity.
    \item \textbf{EulerNet} \cite{EulerNet} leverages Euler’s formula to transform features into the complex vector space, allowing for the efficient modeling high order feature interactions linearly.
    \item \textbf{FiGNN} \cite{FiGNN} represents features as a fully-connected graph and uses gated Graph Neural Networks (GNNs) to model high-order feature interactions.
    \item \textbf{FinalMLP} \cite{FinalMLP} revisits the two–stream paradigm and shows that simply pairing two well-tuned MLPs already yields surprisingly strong CTR results.
    \revision{\item \textbf{SimCEN} \cite{SimCEN} replaces the standard MLP with an alternating explicit–implicit structure enhanced by contrastive learning, where a contrastive product models second-order feature interactions and an external-gated module filters noisy signals.}
\end{itemize}

\subsubsection{Hyper Parameter Setting}
\label{sec: hyper_parameters}
For the sake of fair comparison, the embedding size is uniformly set to 16 across all methods. For each method, we \revision{conducted} grid search over several general hyper parameters to optimize performance. The learning rate is selected from $\{1e^{-3}, 2e^{-3}, 3e^{-3}, 1e^{-4}\}$, while the batch size is varied across $\{512, 1024, 4096\}$. Many methods incorporate the DNN components; we experimented with hidden layer configurations of $\{400 - 400 - 400, 128-128-128, 256-256-256\}$, with a dropout rate of 0.1. The regularization weight penalty is chosen from $\{1e^{-5}, 1e^{-6}\}$. 

We conduct further grid search for tuning the unique hyperparameters associated with each method. In the xDeepFM model, the Compressed Interaction Network (CIN) depth is varied among ${1, 2, 3}$, with the hidden size chosen from $\{100, 200\}$. For DCNV2 model, the low-rank space size is set to 128, while the CrossNet depth is tested across $\{3, 4, 5\}$. For FiGNN model, the number of attention heads is set to 2, and the graph interaction steps are chosen from $\{2, 3, 4\}$. For AFN, the number of logarithmic neurons is selected from $\{400, 800, 1000, 1500\}$. For EulerNet, the number of Euler interaction layers is selected from $\{1, 2, 3\}$, and the number of order vectors is fixed at 30. For FinalMLP, we explored various configurations of DNN hidden layers, as we mentioned above. \revision{For SimCEN, we set the MLP hidden sizes to $[400, 400, 400]$, the contrastive loss weight to $0.1$, and the contrastive temperature to $2.0$.}

For the implicit components of our proposed KarSein model, we choose hidden layers from $\{32-32,64-64,64-32\}$ for all datasets. For the explicit components, we choose the hidden layers from   $\{50-50,26-26,16-16\}$ for the Criteo dataset, and 
$\{4-4,6-6,8-8\}$ for the Douban and MovieLens-1M datasets. Regularization parameters $\lambda_1$ and $\lambda_2$ are selected from $\{1e^{-2}, 1e^{-3}\}$ and $\{1e^{-4}, 1e^{-5}\}$ respectively. The $\kappa$ and $g$ are set from $\{(\kappa=1 , g=3), (\kappa=2 , g=5), (\kappa=3 , g=10) \}$.

\subsubsection{Implementation Details}
We utilize Python 3.8, PyTorch 1.13, CUDA 11.6, and a single Quadro RTX 6000 GPU for implementation. For each method, we reuse the baseline models and implement our models based on RecBole\footnote{https://github.com/RUCAIBox/RecBole}. For evaluation, we employ AUC and LogLoss as metrics to assess the predictive performance of the models. Additionally, we record the number of parameters, excluding those within the embedding layer, to reflect the complexity of each model. \revision{For each method, we run the experiments 10–20 times and report the mean over the top 10 runs. We further perform a two-tailed t-test (p-values) to assess the statistical significance of the improvement of KarSein over the strongest baseline.}

\begin{table*}[h]
\centering
\caption{Performance and parameter size comparison of different models on \textbf{Criteo} and \textbf{Avazu}.}
\label{tab:performance_criteo_avazu}
\begin{tabular}{c|ccc|ccc}
\hline
\multirow{2}{*}{Model}
        & \multicolumn{3}{c|}{Criteo}
        & \multicolumn{3}{c}{Avazu} \\
        & AUC($\uparrow$) & LogLoss($\downarrow$) & Params
        & AUC($\uparrow$) & LogLoss($\downarrow$) & Params \\
\hline
DNN~\cite{DNN_CTR}               & 0.8102 & 0.4417 & 0.226\,M & 0.7918 & 0.3726 & 0.1605\,M \\
Wide \& Deep~\cite{WDL}          & 0.8112 & 0.4409 & 0.226\,M & 0.7920 & 0.3725 & 0.1605\,M \\
DeepFM~\cite{DeepFM}             & 0.8131 & 0.4382 & 0.113\,M & 0.7921 & 0.3723 & 0.080\,M \\
xDeepFM~\cite{xDeepFM}           & 0.8132 & 0.4382 & 0.656\,M & 0.7924 & 0.3722 & 0.3638\,M \\
DCNV2~\cite{DCNv2}               & 0.8135 & 0.4377 & 2.244\,M & 0.7924 & 0.3722 & 1.285\,M \\
AFN+~\cite{AFN}                  & 0.8129 & 0.4397 & 9.982\,M & 0.7921 & 0.3727 & 9.957\,M \\
FiGNN~\cite{FiGNN}               & 0.8130 & 0.4385 & 0.047\,M & 0.7911 & 0.3738 & 0.023\,M \\
EulerNet~\cite{EulerNet}         & 0.8129 & 0.4392 & 0.303\,M & 0.7922 & 0.3723 & 0.179\,M \\
SimCEN~\cite{SimCEN}             & 0.8126 & 0.4396 & 2.414\,M & \underline{0.7937} & \underline{0.3714} & 1.275\,M \\
FinalMLP~\cite{FinalMLP}         & \underline{0.8134} & \underline{0.4378} & 0.739\,M & 0.7929 & 0.3720 & 0.592\,M \\
\hline
KAN~\cite{KAN}                  & 0.8026 & 0.4472 & 0.733\,M & 0.7783 & 0.3864 & 0.5152\,M \\
\textbf{KarSein (Ours)}          & \textbf{0.8145} & \textbf{0.4372} & 0.142\,M & \textbf{0.7951} & \textbf{0.3710} & 0.066\,M \\
\hline
\emph{T-test (p-values)}             &  4.9e-3  &  8.5e-4 &  - &  9.2e-3 &  5.7e-4 & - \\
\hline
\end{tabular}
\end{table*}

\begin{table*}[h]
\centering
\caption{Performance and parameter size comparison of different models on \textbf{Douban Movie} and \textbf{MovieLens-1M}.}
\label{tab:performance_douban_movielens}
\begin{tabular}{c|ccc|ccc}
\hline
\multirow{2}{*}{Model}
        & \multicolumn{3}{c|}{Douban Movie}
        & \multicolumn{3}{c}{MovieLens-1M} \\
        & AUC($\uparrow$) & LogLoss($\downarrow$) & Params
        & AUC($\uparrow$) & LogLoss($\downarrow$) & Params \\
\hline
DNN~\cite{DNN_CTR}               & 0.8305 & 0.3483 & 0.021\,M & 0.8403 & 0.3287 & 0.029\,M \\
Wide \& Deep~\cite{WDL}          & 0.8306 & 0.3482 & 0.021\,M & 0.8410 & 0.3281 & 0.029\,M \\
DeepFM~\cite{DeepFM}             & 0.8307 & 0.3483 & 0.037\,M & 0.8444 & 0.3267 & 0.046\,M \\
xDeepFM~\cite{xDeepFM}           & 0.8292 & 0.3516 & 0.058\,M & 0.8484 & 0.3252 & 0.055\,M \\
DCNV2~\cite{DCNv2}               & 0.8298 & 0.3502 & 0.623\,M & 0.8522 & 0.3192 & 0.697\,M \\
AFN+~\cite{AFN}                  & 0.8300 & 0.3493 & 5.444\,M & 0.8492 & 0.3236 & 5.448\,M \\
FiGNN~\cite{FiGNN}               & 0.8307 & 0.3481 & 0.004\,M & 0.8475 & 0.3243 & 0.006\,M \\
EulerNet~\cite{EulerNet}         & 0.8308 & \underline{0.3480} & 0.253\,M & 0.8531 & 0.3188 & 0.048\,M \\
SimCEN~\cite{SimCEN}             & 0.8312 & 0.3487 & 0.043\,M & 0.8536 & 0.3182 & 0.684\,M \\
FinalMLP~\cite{FinalMLP}         & \underline{0.8316} & 0.3481 & 0.412\,M & \underline{0.8541} & \underline{0.3174} & 0.445\,M \\
\hline
KAN~\cite{KAN}                  & 0.8244 & 0.3657 & 0.053\,M & 0.8273 & 0.3378 & 0.083\,M \\
\textbf{KarSein (Ours)}          & \textbf{0.8335} & \textbf{0.3455} & 0.009\,M & \textbf{0.8555} & \textbf{0.3144} & 0.018\,M \\
\hline
\emph{T-test (p-values)}                  &  2.4e-2 &  2.5e-3 &  -  &   3.9e-3 & 6.6e-4  & - \\
\hline
\end{tabular}
\end{table*}

\subsection{Overall Performance (RQ1)}
\label{sec:overall_performance}
\revision{
This section provides a comparative analysis of the performance and parameter computation cost (excluding the embedding table parameters) between the proposed KarSein model and existing state-of-the-art baselines for CTR prediction. The experimental results are summarized in \autoref{tab:performance_criteo_avazu} and \autoref{tab:performance_douban_movielens}.

The empirical results reveal a stark contrast in performance between KAN and the other models under evaluation. Specifically, KAN consistently underperforms compared to the basic DNN baseline across all four datasets, underscoring its limitations in addressing the demands of CTR prediction tasks. In contrast, our proposed method, KarSein, demonstrates robust and consistent superiority, outperforming all baseline approaches across the same datasets. KarSein achieves statistically significant improvements in AUC over strong state-of-the-art models such as EulerNet and FinalMLP, classical baselines like DCNv2, as well as the recently proposed contrastive-learning method SimCEN. On Criteo and Avazu, the AUC gains over these baselines exceed $0.001$ (i.e., $0.1\%$). \textbf{Typically, an improvement of $0.1\%$ in AUC is considered significant for CTR tasks \cite{chen2021enhancing,DCNv2,BARS,light_weight_benchmark}.} Consistently, the paired $t$-tests reported in the last row of \autoref{tab:performance_criteo_avazu} and \autoref{tab:performance_douban_movielens} yield $p$-values below $0.01$, confirming that these gains are statistically reliable. This notable performance gain indicates that KarSein's success stems from the synergistic integration of its key components, particularly the learnable activation function mechanism, which plays a pivotal role in effectively capturing intricate high-order feature interactions. 

From the perspective of computational cost, models such as KAN, FinalMLP, AFN+, DCNv2, xDeepFM, and SimCEN exhibit substantially larger parameter sizes. In contrast, the KarSein model demonstrates a markedly smaller parameter size.
Across the four datasets, KarSein uses between roughly $2\times$ and over $600\times$ fewer parameters than strong baselines, and about $5$–$8\times$ fewer parameters than KAN, while consistently achieving higher AUC. This empirical observation aligns with our prior theoretical analysis of FLOPs, further validating the efficiency-oriented design principles of KarSein.

}

\subsection{Ablation Study (RQ2)}

We conducted a comprehensive investigation into the contributions of explicit and implicit feature interactions within the KarSein model by systematically isolating and analyzing each component. Specifically, we decomposed the default ensemble KarSein model, which integrates both interaction types, to evaluate the performance, efficiency, and computational complexity associated with each type independently. This analysis encompassed metrics such as CTR prediction accuracy, model size, and training time. Our results, summarized in \autoref{tab:vec_bit_ablation_a} and \autoref{tab:vec_bit_ablation_b}, reveal that the explicit interaction component is the primary driver of performance improvements. Notably, the model leveraging only explicit interactions achieves an AUC performance comparable to the full ensemble model while offering significant reductions in model size and training time. This efficiency underscores the capacity of explicit interactions to capture essential feature dependencies with minimal computational overhead. Conversely, the implicit interaction component, while contributing to marginal AUC gains when combined with explicit interactions, exhibits diminished standalone effectiveness. 

These findings underscore the complementary nature of explicit and implicit interactions. While the explicit component excels in efficiency and parameter economy, the optional inclusion of implicit interactions can serve as an auxiliary mechanism to further optimize predictive accuracy in scenarios where resource constraints are less critical.

\begin{table*}[h]
\centering
\caption{Impact of Feature Interaction Mechanisms on KarSein Model: Criteo and Avazu.}
\label{tab:vec_bit_ablation_a}
\begin{tabular}{c|ccc|ccc}
\hline
\multirow{2}{*}{Model} & \multicolumn{3}{c|}{Criteo} & \multicolumn{3}{c}{Avazu} \\
                       & AUC & Params & Time×Epochs & AUC & Params & Time×Epochs \\
\hline
KarSein-explicit (vector-wise)       & 0.8142 & 0.051 K & 30m~× 3  & 0.7944 & 0.013 M & 24.68m x 2 \\
KarSein-implicit (bit-wise)       & 0.8125 & 0.091 M & 23m~× 4  & 0.7912 & 0.053 M & 24.78m x 2 \\
KarSein                & 0.8145 & 0.092 M & 40m~× 3  & 0.7951 & 0.066 M & 27.9m x 5 \\
\hline
\end{tabular}
\vspace{-3mm}
\end{table*}

\begin{table*}[h]
\centering
\caption{Impact of Feature Interaction Mechanisms on KarSein Model: Douban Movie and MovieLens-1M.}
\label{tab:vec_bit_ablation_b}
\begin{tabular}{c|ccc|ccc}
\hline
\multirow{2}{*}{Model} & \multicolumn{3}{c|}{Douban Movie} & \multicolumn{3}{c}{MovieLens-1M} \\
                       & AUC & Params & Time×Epochs        & AUC & Params & Time×Epochs \\
\hline
KarSein-explicit (vector-wise)       & 0.8319 & 0.269 K & 14.4s~× 3  & 0.8542 & 0.812 K & 8.4s~× 3  \\
KarSein-implicit (bit-wise)       & 0.8311 & 0.008 M & 13.2s~× 2  & 0.8535 & 0.017 M & 5.7s~× 10 \\
KarSein                & 0.8335 & 0.009 M & 19.5s~× 2  & 0.8555 & 0.018 M & 9.9s~× 3  \\
\hline
\end{tabular}
\vspace{-3mm}
\end{table*}

\subsection{Impact of Hyperparameters (RQ3)}
\label{sec: hyper_param_exp}
In this section, we study how three key groups of hyperparameters affect KarSein’s performance: (i) the configuration of pairwise multiplication layers, (ii) the network width and depth, and (iii) the spline activation parameters.

\begin{figure}[h]
  \centering
    \includegraphics[width=3.8 in]{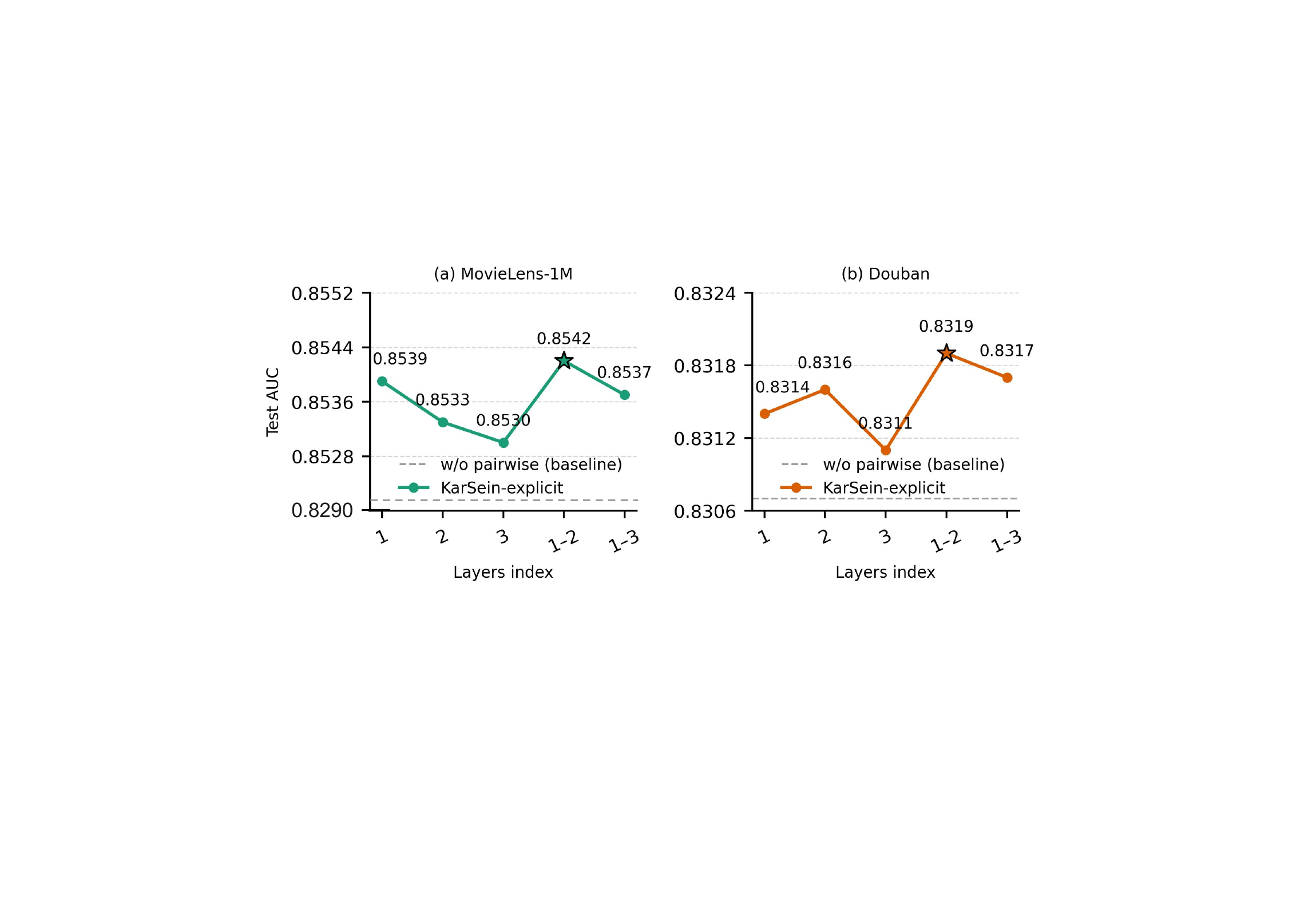}
   \caption{Impact of Pairwise Multiplication in Different Configurations on AUC Performance of the KarSein-explicit}
    \label{fig: feature_crosses_impact}
\end{figure}

\subsubsection{Impact of Pairwise Multiplication}
To investigate how the configuration of pairwise multiplication affects the model's performance, we analyze the impact of applying this operation across different layers. Specifically, we empirically demonstrate that restricting pairwise multiplication to only the first two interaction layers is sufficient for the model to effectively learn multiplicative relationships. \autoref{fig: feature_crosses_impact} presents the performance comparison of KarSein-explicit on the MovieLens-1M and Douban datasets under various configurations of pairwise multiplication layers. When pairwise multiplication is omitted (Layer Index set to None), the model's AUC performance is notably poor. Introducing this step in just the first layer leads to significant performance gains, especially on the MovieLens-1M dataset. We attribute this to the presence of six fundamental field features in the MovieLens-1M dataset, which provide a rich diversity of learnable multiplicative interactions. In contrast, the Douban dataset, with only two features, exhibits comparatively limited improvement.

These empirical findings corroborate our earlier assertion that the KAN network struggles to capture multiplicative relationships, thereby constraining its effectiveness in CTR prediction tasks. They also validate the efficacy of a straightforward mechanism: applying pairwise multiplication in the initial interaction layers significantly enhances KarSein's ability to learn multiplicative feature interactions. Besides, according to our empirical experience, we recommend that configuring pairwise multiplication within the first two layers can yield the highest AUC scores.

\begin{figure}[h]
  \centering
    \includegraphics[width=3.8 in]{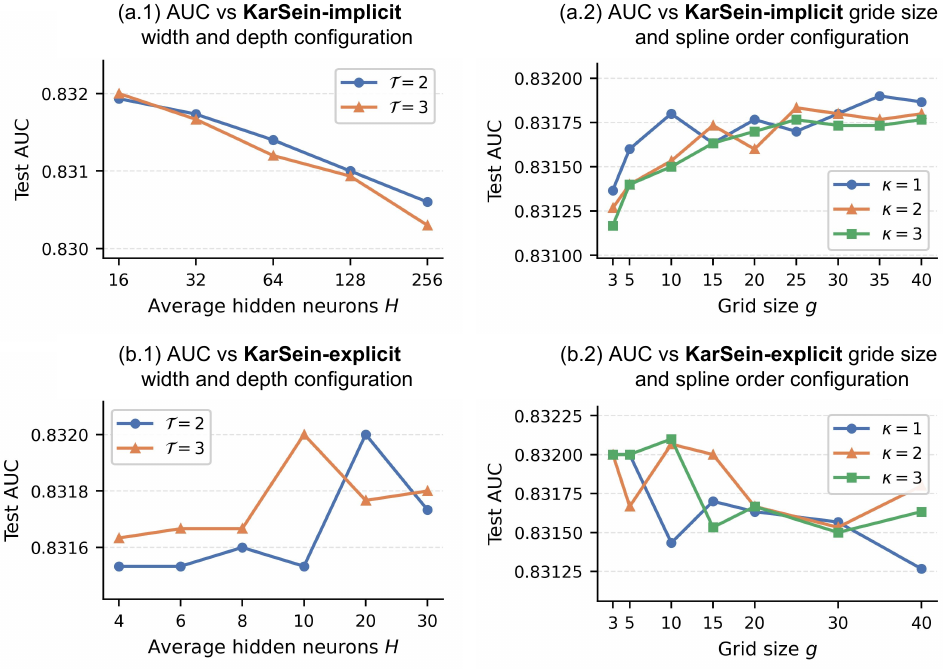}
   \caption{\revision{Hyperparameter study of KarSein.
Panels (a.1)–(a.2) report the test AUC of KarSein-implicit when varying the width/depth (average hidden neurons $H$ and number of hidden layers $\mathcal{T}$) and the spline parameters (grid size $g$ and order $\kappa$); panels (b.1)–(b.2) show the corresponding results for KarSein-explicit.}}
    \label{fig: scaling}
\end{figure}

\revision{
\subsubsection{Impact of Network Width and Depth}
We conduct this investigation on the Douban dataset. We fix the pairwise multiplication operations to layers 1–2. Regarding spline activation configurations, we set $(\kappa=2,g=5)$ for KarSein-implicit and $(\kappa=3,g=10)$ for KarSein-explicit. We then vary the hidden width $H$ and depth $\mathcal{T}$ to investigate the sensitivity of both branches to architectural changes; the results are presented in \autoref{fig: scaling}~(a.1) and (b.1).
For KarSein-implicit, enlarging $H$ from 16 to 256 leads to a consistent decrease in AUC, and the deeper setting $\mathcal{T}=3$ does not outperform $\mathcal{T}=2$.  
This suggests that scaling the implicit branch increases parameters without yielding additional predictive power.  
In contrast, the explicit branch shows a different pattern and can benefit from expanding width: as $H$ increases from 4 to 30, AUC first improves and then slightly degrades, with the best performance around mid-range widths (e.g., $H\approx 10$–20). Overall, the deeper configuration ($\mathcal{T}=3$) tends to match or slightly outperform $\mathcal{T}=2$, but at the cost of additional computation, and a carefully tuned shallow network with moderate width can also saturate the performance of this branch.

Taken together, these observations suggest a small-but-strong regime: near-optimal performance is already achieved with relatively narrow and shallow KarSein branches, which is appealing for latency- and memory-sensitive CTR systems. In practice, this allows practitioners to restrict the search over $(H,\mathcal{T})$ to a compact region—for example, as our Douban experiments indicate, using small $H$ for KarSein-implicit and $m \leq H \leq m^2$ with $\mathcal{T}\in{2,3}$ for KarSein-explicit—thereby reducing hyperparameter tuning cost while preserving state-of-the-art accuracy.

\subsubsection{Impact of Spline Activation}
Continuing our investigation on the Douban dataset, we fix the network architecture to $(H=32, \mathcal{T}=2)$ for KarSein-implicit and $(H=10, \mathcal{T}=3)$ for KarSein-explicit. We systematically vary the spline activation hyperparameters, the grid size $g$ and spline order $\kappa$ for analyzing their impact on model performance. The results are illustrated in \autoref{fig: scaling}~(a.2) and (b.2).
For the implicit branch, increasing $g$ from 3 to around 15 steadily improves AUC, after which the curve flattens or has small improvement with fluctuations; different orders $\kappa\in{1,2,3}$ yield almost overlapping trajectories. This indicates that a moderate grid size cooperated with a suitable spline order is sufficient for the implicit spline to capture useful non-linearities, and higher-order splines do not provide clear benefits for CTR signals. For the explicit branch, AUC changes little when $g$ increases from small to medium values, and then gradually decreases as the grid becomes finer. Likewise, higher spline orders do not bring consistent gains: the best performance is obtained with small-to-medium grids combined with a suitable $\kappa$.

These trends imply that KarSein does not rely on aggressively over-parameterized spline bases; simple low-order activations with moderate grid sizes already provide enough functional flexibility. From a tuning perspective, one can fix $\kappa$ to a small value (e.g., $\kappa=2$) and select $g$ from a narrow range (e.g. $\leq 15$), achieving a good trade-off between accuracy and parameter efficiency.
}

\subsection{Explanation Study (RQ4)} \label{sec:explanatin}

\begin{figure}[ht]
  \centering
    \includegraphics[width=3.0 in]{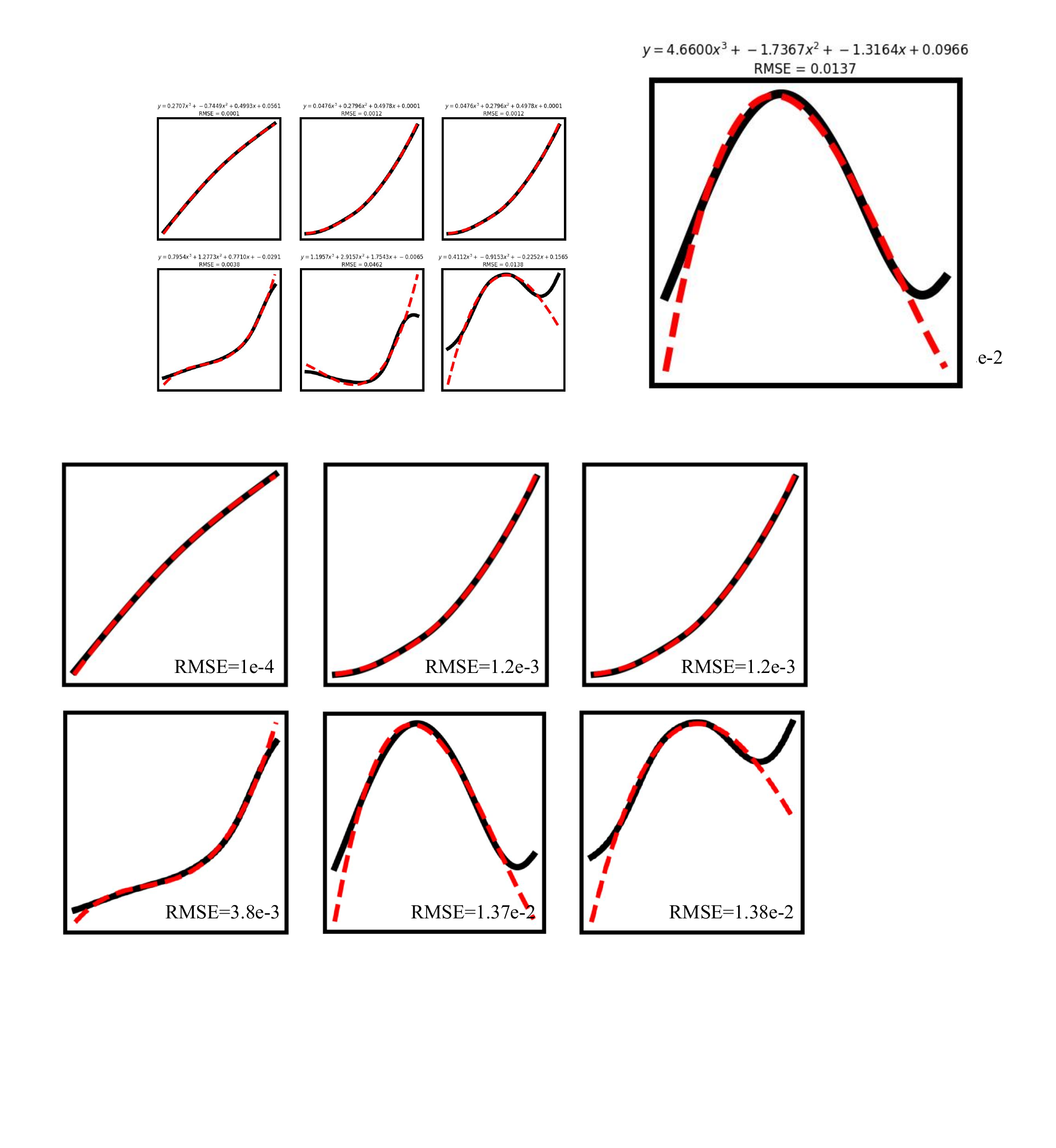}
   \caption{Visualization of several representative learnable activation functions in KarSein with $\kappa=3$, $g=10$. As an example, the activation function in the top-left corner has a symbolic regression expression of $y = 0.2707x^3 - 0.7449x^2 + 0.4993x + 0.0561$, achieving a fitting performance with RMSE = $1 \times 10^{-4}$.}
    \label{fig: activation_visualiz}
\end{figure}

Our further exploration of the KarSein model focuses on how learnable activation functions transform low-order input features into output features. We visualized these activation functions across various layers and used third-degree polynomials for symbolic regression. Most activation functions were well-approximated by cubic polynomials, as shown in the first row of \autoref{fig: activation_visualiz}. This indicates that KarSein's activation functions effectively elevate low-order features to capture high-order interactions.

\revision{We also observe a subset of activation functions exhibiting oscillatory or highly irregular patterns (second row of \autoref{fig: activation_visualiz}). These shapes cannot be reliably captured by cubic polynomials and typically require quartic, quintic, or even higher-degree expressions in symbolic regression. This behavior highlights the expressive power of B-spline activations: they provide sufficient flexibility to model very complex, high-order nonlinearities when needed, which in turn contributes to the strong empirical performance of KarSein.
}

\revision{
According to the above empirical findings, we now explicitly characterize the interaction order realized by the KarSein-explicit branch as a function of network depth. We continue to use the notation introduced in \autoref{sec:karsein_interaction_layer}. Let $\mathcal{T}$ denote the depth (number of stacked KarSein Interaction Layers) in the explicit branch, and let $H_L$ be the number of hidden neurons in the $L$-th layer, with $H_0 = m$ and $H_{\mathcal{T}} = 1$. For brevity, we also write $H$ for a typical hidden width, which in practice is chosen in the range $H \in [m, m^2]$.
As described in \autoref{sec:pairwise_multiplication}, the pairwise-multiplication operator $\mathcal{P}_L$ takes an input representation of interaction order $O_{L-1}$ and forms pairwise products with the original fields whose order is 1. This increases the degree by one, so the order after $\mathcal{P}_L$ is
\[
O_{L-1}
\;\gets\;
O_{L-1} + \mathbf{1}\{\mathcal{P}_L \text{ is applied}\},
\]
where $\mathbf{1}\{\cdot\}$ is the indicator function. For example, $\mathbf{X}^0$ only contains first-order terms, and in our explicit branch, we activate $\mathcal{P}_1$ and $\mathcal{P}_2$ at the entrance of the first two KarSein layers. Concretely,
\[
O_0
= 1 + \mathbf{1}\{\mathcal{P}_1 \text{ is applied}\}
= 2,
\qquad
O_1
\;\gets\;
O_1 + \mathbf{1}\{\mathcal{P}_2 \text{ is applied}\}.
\]

Under the empirical findings that each learned activation is well approximated by a univariate polynomial of degree $r \ge 3$, each KarSein Interaction Layer $\Psi_L = \mathcal{L}_L \circ \mathcal{A}_L \circ \mathcal{P}_L$ takes features of maximum interaction order $O_{L-1}$ (after the possible update by $\mathcal{P}_L$) and applies degree-$r$ univariate transformations followed by linear mixing. This multiplies the maximum interaction order by $r$ at each depth, so the maximum interaction order after layer $L$ is
\[
O_L = r\,O_{L-1}, \qquad L = 1,\dots,\mathcal{T},
\]
Unrolling the recursion for the configuration used in our experiments, where $\mathcal{P}_1$ and $\mathcal{P}_2$ are activated and $\mathcal{P}_L$ is turned off for $L \ge 3$, yields
\[
O_1 = r\bigl(1 + \mathbf{1}\{\mathcal{P}_1 \text{ is applied}\}\bigr)
     = r(1+1)
     = 2r,
\]
\[
O_2 = r\bigl(O_1 + \mathbf{1}\{\mathcal{P}_2 \text{ is applied}\}\bigr)
     = r(2r+1).
\]
\[
O_3 = r\bigl(O_2 + 0\bigr)
     = r(2r^2 + r)
     = 2r^3 + r^2.
\]
Since in all our KarSein-explicit configurations we use $\mathcal{T}=3$ interaction layers (i.e., architectures of the form $m$–$H$–$H$–1), the maximum interaction order at the output of the explicit branch is therefore
\begin{equation}
\label{eq:order_bound_general}
O_{\mathcal{T}} = O_3 = 2r^3 + r^2.
\end{equation}
The final prediction head on top of $\mathbf{X}^{\mathcal{T}}$ is linear in the explicit features and thus does not further change this order. For $r \ge 3$, \eqref{eq:order_bound_general} implies
\[
O_{\mathcal{T}}
= 2r^3 + r^2
\;\ge\;
2\cdot 3^3 + 3^2
= 2\cdot 27 + 9
= 63,
\]
showing that KarSein-explicit can realize interactions of at least 63rd order in all these settings.

More generally, this analysis gives a simple rule of thumb for designing KarSein-explicit. Let $O_0 = 1$ denote the order of original field features, choose an activation degree $r$, a depth $\mathcal{T}$, and an on/off schedule for pairwise-multiplication layers encoded by indicators $s_L \in \{0,1\}$, where $s_L = 1$ if $\mathcal{P}_L$ is applied and $s_L = 0$ otherwise. Then the maximum interaction order follows the recursion
\[
O_L = r\bigl(O_{L-1} + s_L\bigr), \qquad L = 1,\dots,\mathcal{T},
\]
which admits the closed-form expression
\begin{equation}
\label{eq:order_calculation}
O_{\mathcal{T}}
= r^{\mathcal{T}}
  + \sum_{L=1}^{\mathcal{T}} s_L\, r^{\mathcal{T}-L+1}
=O(r^{\mathcal{T}}).
\end{equation}
In our experiments, setting $s_1 = s_2 = 1$ and $s_L = 0$ for $L \ge 3$ recovers \autoref{eq:order_bound_general}.
}

\subsection{Structural Sparsity Observation (RQ5)}
\begin{figure}[t]
  \centering
    \includegraphics[width=5.2 in]{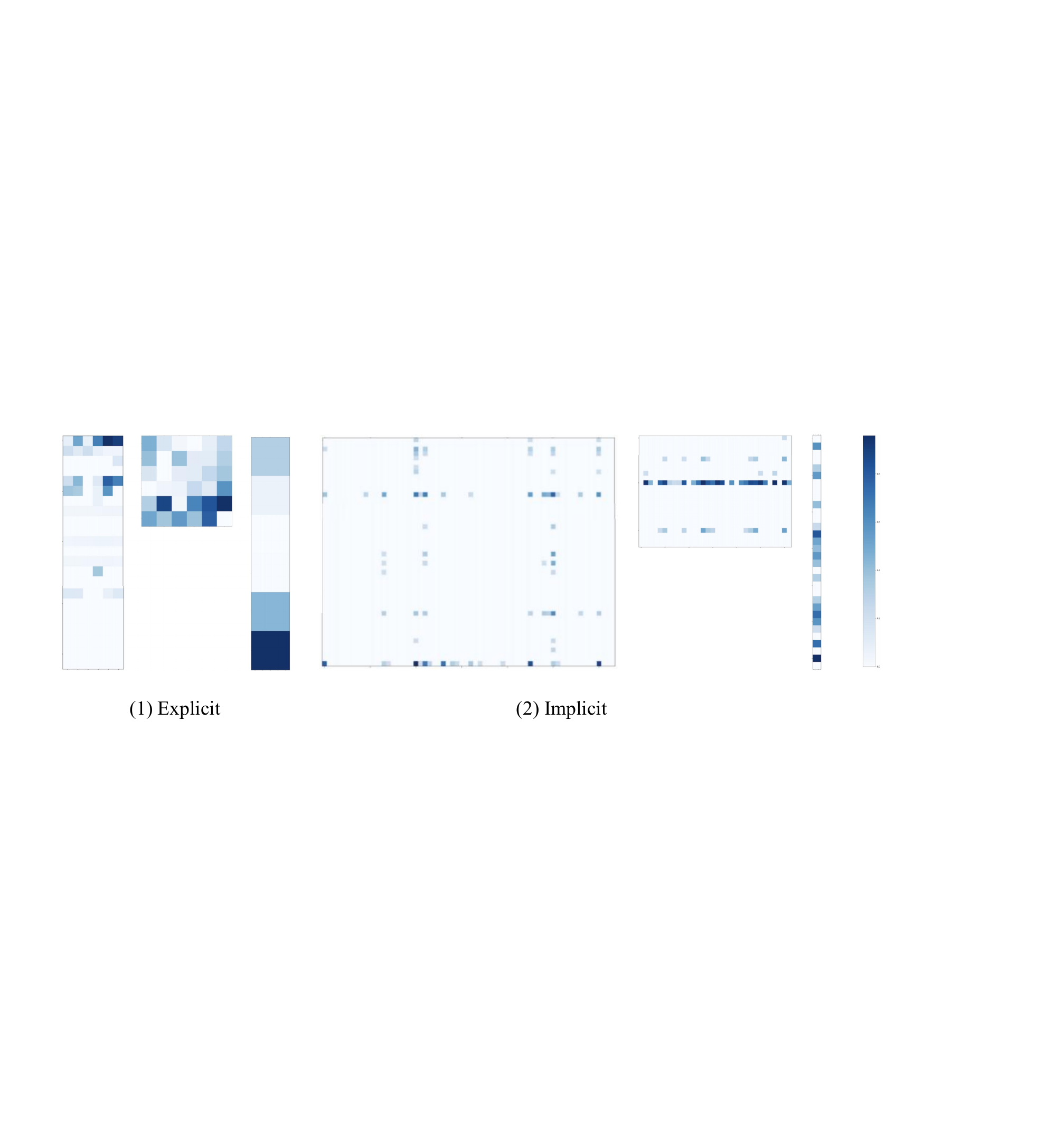}
   \caption{Layer-Wise Visualization of Feature Connections in the $\text{KarSein}$ Model. Subfigure (1) focuses on the explicit component, where $D=16$ matrices correspond to 16 different dimensions of embedding vectors. For simplicity and clarity, we average these matrices and visualize the result as a heatmap. Subfigure (2) represents the implicit component. Each row in the heatmap corresponds to an input feature, and each column to an output feature. The color intensity reflects the connection weights. To enhance the aesthetic and interpretability of the visualization, most of the blank areas in the heatmaps have been removed.}
    \label{fig: sparsity}
\end{figure}

To assess the structural sparsity of the optimized KarSein model, we conducted an investigation using the KarSein model fine-tuned on the MovieLens-1M dataset, which achieved an AUC of 0.8555. Heat maps were employed to visualize the connections between activated inputs and outputs across each layer of the model, as depicted in \autoref{fig: sparsity}.

Our analysis revealed that a substantial proportion of input-output edge connection weights were $\leq 0.01$. This observation highlights the efficacy of the regularization term in KarSein. Consequently, the converged KarSein model exhibits a highly sparse and efficient network structure. This inherent sparsity endows KarSein with the capability to prune the network into a more compact subnetwork, similar to the pruning capabilities of KAN.

Building on this insight, we further masked the negligible edge connections and continued tuning the KarSein model for an additional three epochs until convergence. The resulting model achieved an AUC of 0.8533. Despite already having a minimal parameter size, KarSein demonstrated potential for further compression while retaining competitive predictive performance. This underscores the promise of highly sparse and efficient network architectures.

\section{Related Works}
\subsection{Feature Interaction Types}

\revision{
Click-Through Rate (CTR) prediction is a key component of modern recommender systems, crucial for accurate content delivery and user engagement \cite{lee2007context,liu2010personalized,xu2024slmrec,xu2023neural,xu2024rethinking,huang2023modeling}. Two broad paradigms exist for modeling feature interactions in CTR prediction. The first, the \emph{implicit} paradigm, often employs deep neural network (DNN) architectures, such as MLP-based stacks and hybrid towers \cite{DNN_CTR,WDL,FinalMLP,NCF,qu2016product}, as well as sequential or Transformer-style encoders \cite{FuXi,LSTM_ctr,huang2023dual2}. Underpinned by the Universal Approximation Theorem, these models can capture latent interactions and complex nonlinear patterns. However, their interpretability remains limited because interactions are implicitly entangled within deep hidden layers. Moreover, they are inherently less effective at modeling multiplicative relationships—an aspect shown to be crucial for CTR prediction and challenging for standard implicit models to learn without architectural bias \cite{AutoInt,rendle2020neural}. By contrast, the \emph{explicit} paradigm constructs high-order feature combinations by performing inner products among basic feature embedding vectors, typically in an enumerative manner. These interactions can be expressed as higher-order polynomials of the input features and are commonly implemented through factorization or polynomial-style modules. Representative methods include FwFMs, NFM, and DeepFM \cite{FwFMs,NFM,DeepFM}, the Compressed Interaction Network (CIN) in xDeepFM \cite{xDeepFM}, large-scale industrial frameworks such as DLRM \cite{DLRM}, and higher-order variants \cite{High_Order_User_Preference,DHFM}. In practice, many CTR systems therefore adopt a hybrid design: explicit interaction modules provide strong multiplicative signals, and implicit interactions serve as a complementary signal to explicit components, an ensemble pattern widely instantiated in WDL, DLRM, DeepFM, xDeepFM, DeepIM, DCNv2 \cite{WDL,DLRM,DeepFM,xDeepFM,DeepIM,DCNv2, SimCEN}.
}

\subsection{Adaptive Order Feature Interaction Learning}

\revision{
In explicit feature-interaction modeling, recent advances pursue high-order feature interactions without manually enumerating cross terms or predefining the order. For example, AFN performs linear mixing in log space so that exponentiation emulates multiplicative interactions \cite{AFN,LNN}, whereas EulerNet maps embeddings into a geometric/polar domain to synthesize interactions before projecting back \cite{EulerNet}. These methods reduce the combinatorial overhead while effectively enabling models to capture more complex patterns and often improve predictive performance. However, they rely on cross-space reparameterizations with numeric assumptions (e.g., positivity for logarithms) and additional forward/backward mappings, which introduce numerical stability issues and additionally computational overhead. It may also complicate optimization and obscure the semantics of interactions in the original feature space. KarSein instead directly activates features within the original space to adaptive orders using a learnable activation function. Our approach is much simpler and more effective, also has intuitive symbolic regression explanations.
}

\subsection{Feature Importance Learning}

\revision{
The number of candidate feature interactions grows exponentially with order, so many methods not only model higher-order interactions but also select the most salient ones \cite{Beneficial,Arbitrary_Order_Beneficial}. AFM emphasizes pairwise terms \cite{AFM}; AutoInt applies multi-head self-attention across fields \cite{AutoInt}; FiGNN treats fields as nodes in an interaction graph \cite{FiGNN}; and FiBiNET augments bilinear functions with channel-wise reweighting \cite{FiBiNET,SENET,attention}. These mechanisms yield input-dependent saliency and often improve accuracy, yet their scores are instance-specific, making global, structure-level interpretation unstable. In contrast, KarSein enforces a sparse, globally consistent interaction structure via sparsity regularization, so the same edges and composing functions carry meaning across inputs. This preserves the benefits of multiplicative modeling while providing a compact, dataset-level view of which higher-order interactions the model truly relies on. By decoupling interpretability from per-sample attention weights and imposing structure that is independent of the recommendation context, KarSein offers a unified, globally interpretable notion of feature importance.
}

\section{Conclusion and Future Work}

\revision{
In this paper, we leveraged the learnable activation mechanism of Kolmogorov–Arnold Networks (KANs) and proposed KarSein, a Kolmogorov–Arnold Represented Sparse-Efficient Interaction Network tailored for CTR prediction. It injects lightweight pairwise-multiplication “catalysts’’ in the early layers to make multiplicative patterns easy to discover, and replaces KAN’s redundant multi-activation design with a single spline-based activation per feature equipped with sparsity regularization. At the architectural level, KarSein extends KAN from scalar inputs to vector-wise feature interactions, allowing embedding vectors to be treated as holistic units rather than independent coordinates. These design choices directly address the three key limitations we identified in vanilla KAN for CTR—its difficulty in discovering multiplicative relations, its computational redundancy, and its incompatibility with embedding-based vector-wise interactions—while preserving KAN’s strengths in interpretability and structural sparsity.

Extensive experiments on four real-world datasets (Criteo, Avazu, MovieLens-1M, and Douban Movie) demonstrate that KarSein achieves state-of-the-art CTR prediction accuracy with a fraction of the parameter cost of strong baselines. KarSein consistently outperforms both classical models (e.g., DCNv2, DeepFM, xDeepFM) and more recent architectures such as EulerNet, FinalMLP, and SimCEN. Importantly, it does so while using, on average, around $50\times$ fewer parameters than strong baselines. Ablation studies further reveal that the explicit (vector-wise) branch is the primary driver of these gains, while the implicit (bit-wise) branch provides complementary improvements when latency and memory budgets allow. Beyond aggregate performance, our analyses provide a more fine-grained understanding of how KarSein behaves and why it is effective. Hyperparameter studies show that KarSein operates in a “small-but-strong’’ regime: near-optimal performance is already achieved with relatively narrow and shallow interaction branches, and moderate spline grid sizes with low spline orders are sufficient. This yields concrete tuning guidelines for practitioners deploying KarSein in production CTR systems. The explanation study demonstrates that the learned spline activations can be well approximated by low-degree polynomials and, through symbolic regression, reveals that KarSein’s explicit branch naturally realises very high-order feature interactions—with at least 63rd-order interactions in the 3-layer configurations used in our experiments. Finally, the structural sparsity study shows that a large fraction of edges in the learned network carry negligible weights, enabling further pruning with minimal loss in accuracy and offering a globally interpretable view of which high-order interactions the model truly relies on.

In summary, this work makes KarSein a practical, parameter-efficient, and theoretically grounded alternative to existing CTR models, capable of delivering state-of-the-art accuracy together with high-order interpretability and compact, deployable architectures.
}

\begin{acks}
This work was sponsored by the \texttt{Australian Research Council under the Linkage Projects Grant LP210100129}.
\end{acks}

\bibliographystyle{ACM-Reference-Format}
\bibliography{acmart}

\end{document}